\pdfoutput=1

\documentclass[11pt]{article}

\usepackage[final]{acl}

\usepackage{times}
\usepackage{latexsym}

\usepackage[T1]{fontenc}

\usepackage[utf8]{inputenc}

\usepackage{microtype}

\usepackage{inconsolata}

\usepackage{graphicx}
\usepackage{amsmath}
\usepackage{bm}

\usepackage{times}
\usepackage{enumitem} 
\usepackage{float}

\usepackage{array}
\usepackage{tabularx}
\usepackage{booktabs}
\usepackage{booktabs}
\usepackage{xcolor}
\newcolumntype{Y}{>{\arraybackslash}X}

\usepackage{graphicx}
\usepackage{subcaption}

\title{Hop, Skip, and Overthink: Diagnosing Why Reasoning Models Fumble during Multi-Hop Analysis}

\author{
 \textbf{Anushka Yadav\textsuperscript{2*}},
  \textbf{Isha Nalawade\textsuperscript{2*}},
  \textbf{Srujana Pillarichety\textsuperscript{2*}},
  \textbf{Yashwanth Babu\textsuperscript{2*}},
\\
  \textbf{Reshmi Ghosh\textsuperscript{1}},
  \textbf{Samyadeep Basu\textsuperscript{3}},
  \textbf{Wenlong Zhao\textsuperscript{2}},
\\
  \textbf{Ali Nasaeh\textsuperscript{2}},
  \textbf{Sriram Balasubramaniam\textsuperscript{3}},
\textbf{Soundararajan Srinivasan\textsuperscript{1}}
  \\
\small{* equal contribution}
\\
  \textsuperscript{1}Microsoft,
 \textsuperscript{2}University of Massachusetts, Amherst,
  \textsuperscript{3}University of Maryland, College Park,
\\
 \small{
    \textbf{Correspondence:} \href{mailto:reshmighosh@microsoft.com}{reshmighosh@microsoft.com}
  }
}

\begin{document}
\maketitle
\begin{abstract}
The emergence of reasoning models and their integration into practical AI chat bots has led to breakthroughs in solving advanced math, deep search, and extractive question answering problems that requires a complex and multi-step thought process. Yet, a complete understanding of why these models hallucinate more than general purpose language models is missing.
In this investigative study, we systematically explore reasoning failures of contemporary language models on multi-hop question answering tasks. We introduce a novel, nuanced error categorization framework that examines failures across three critical dimensions: the diversity and uniqueness of source documents involved ("hops"), completeness in capturing relevant information ("coverage"), and cognitive inefficiency ("overthinking"). Through rigorous human annotation, supported by complementary automated metrics, our exploration uncovers intricate error patterns often hidden by accuracy-centric evaluations. This investigative approach provides deeper insights into the cognitive limitations of current models and offers actionable guidance toward enhancing reasoning fidelity, transparency, and robustness in future language modeling efforts.
\end{abstract}

\section{Introduction}

Language models (LMs) have demonstrated remarkable performance on multi-hop question answering (QA) benchmarks, such as HotpotQA \cite{hotpotqa}, where success requires sourcing knowledge from multiple documents. 
MuSiQue \cite{musique} extends this task by posing harder questions that reduces shortcut reasoning and provides explicit reasoning paths to better assess multi-step inference.


The traditional evaluation metrics employed in these tasks, such as the final answer accuracy or the F1 score, fail to distinguish between genuine multi-step inference, simple memorization (as exposed by counterfactual benchmarks such as CofCA; \cite{wu2025cofca}, and over-reliance on dataset artifacts. Moreover, emerging studies \cite{interpretinglanguagemodelscase, faithfulnessvsplausibilityunreliability} show that errors may stem from missing knowledge recall, misinterpretation of question intent, or retrieval failures in retrieval-augmented settings. 

\begin{figure}
    \centering
    \includegraphics[width=0.8\linewidth]{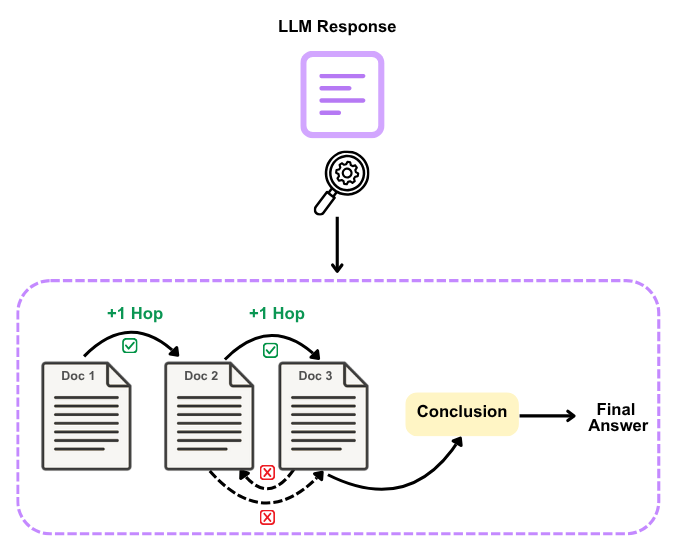}
    \caption{
\textbf{Illustration of Multi-hop Reasoning}
A language model answers via a three-hop reasoning chain. The back-and-forth between documents illustrates overthinking, where unnecessary steps are added beyond the ideal inference path.
}
    \label{fig:hop-transition}
\end{figure}

With these limitations in mind, we move beyond answer correctness and undertake an investigative exploration of reasoning failures in multi-hop QA to answer a central question: How and why do reasoning models break down when stitching together information across multiple sources? 
To address this,  we introduce a diagnostic framework that decomposes reasoning behavior along three core dimensions:
(1) \textbf{Hops} A hop is a discrete step or transition in the reasoning process where the model moves from one piece of information (e.g., a fact, source, or knowledge base entry) to another in order to bridge connections and form a complete answer.
(2) \textbf{Coverage} evaluates whether all necessary reasoning steps are covered; and
(3) \textbf{Overthinking} refers to whether the model meanders into unnecessary or off-track reasoning.
These dimensions support both qualitative annotation and targeted quantitative evaluation of reasoning fidelity.




In this paper, we introduce a detailed set of reasoning error categories and apply them to manually annotate model traces from six language models across three multi-hop QA datasets. Building on these annotations, we develop an automated evaluation framework that closely mirrors human judgments.


In this paper, we define seven fine-grained reasoning error categories and manually annotate up to 80 responses across three datasets and six models. To scale this analysis, we introduce an LLM-as-a-Judge framework that achieves 74\% hop match accuracy and 50–75\% label agreement with human annotations on 2Wiki, MuSiQue, and HotpotQA.

\paragraph{Contributions:}
We present a detailed analysis of multi-hop reasoning by curating and annotating model responses on three diverse datasets: 2WikiMultiHopQA, HotpotQA, and MuSiQue. Using a structured error taxonomy, we quantify the distribution of reasoning errors across models. Our study reveals common reasoning issues, such as breaking down in the middle of reasoning, adding unnecessary steps in complex cases, and providing correct answers despite flawed reasoning, especially on questions with many entities or confusing information. Finally, we evaluate the effectiveness of an LLM-as-a-Judge framework, which shows strong agreement with human annotations on simpler datasets while highlighting key limitations on more complex ones. This supports the use of scalable, semi-automated evaluation for reasoning analysis.






\section{Related Works}

Reasoning in large language models has progressed from relying on statistical correlations to adopting more structured mechanisms like Chain-of-Thought (CoT) prompting \cite{wei2023chainofthoughtpromptingelicitsreasoning}. Despite this, several studies have shown that LLMs often generate unfaithful explanations that do not reflect the true reasoning path, with causal analyses suggesting that these traces are frequently post hoc rationalizations \cite{bao2024likelyllmscotmimic}. Recent work on Large Reasoning Models finds that, while they can outperform standard LLMs on medium-complexity tasks, they exhibit surprising scaling limits and collapse in accuracy on high-complexity problems despite detailed reasoning traces \cite{shojaee2025illusionthinkingunderstandingstrengths}. Traditional metrics like F1 and BLEU focus on answer correctness but overlook reasoning quality, and multihop QA benchmarks have revealed that models often exploit shortcuts to arrive at correct answers without faithfully connecting supporting evidence \cite{ishii-etal-2024-analysis}. Heuristic-based evaluations can further mask such reasoning failures \cite{lanham2023measuringfaithfulnesschainofthoughtreasoning}, prompting the need for more targeted reasoning assessments.

To address this, recent work has focused on identifying and analyzing intermediate reasoning errors. Studies show that correcting flawed steps can improve model robustness \cite{li2024evaluatingmathematicalreasoninglarge}, while ProcessBench highlights issues like process errors and logical inconsistency during multi-step reasoning \cite{zheng2024processbenchidentifyingprocesserrors}. Adding explicit premises to reasoning chains has been shown to improve error detection and clarity \cite{mukherjee2025premiseaugmentedreasoningchainsimprove}. However, challenges remain: hallucinations and factual inconsistencies in long-form outputs are still hard to detect even with strong models \cite{kamoi2024evaluatingllmsdetectingerrors}, and repeated reasoning mistakes persist without explicit supervision \cite{tong2024llmslearnpreviousmistakes}.

Beyond standard QA settings, more recent evaluations have looked at reasoning in complex domains. Work on Olympiad-style math problems finds that models often produce shallow or incomplete reasoning despite correct final answers \cite{mahdavi2025brainsvsbytesevaluating}. Similarly, in multimodal settings, ErrorRadar benchmarks expose systematic reasoning failures in math-heavy questions, reinforcing the importance of fine-grained reasoning analysis beyond surface-level correctness \cite{yan2024errorradarbenchmarkingcomplexmathematical}. Our work builds on these insights by explicitly annotating multi-hop reasoning traces and categorizing failure patterns across diverse QA datasets, enabling a more principled and scalable evaluation of reasoning quality.

\section{Methodology}


 


\subsection{Task Formalization}
We define \textbf{multi-hop QA} as the task of responding to complex questions, undertaken by reasoning models, that necessitate synthesizing information from multiple sources through a chain of reasoning steps. A \textbf{hop}, denoted by \( h_i \), refers to a distinct reasoning step wherein the model extracts supporting evidence from a \textbf{unique document} \( d_j \in D \). The number of hops in a reasoning path corresponds to the number of unique documents accessed, regardless of how much content is extracted from each. Figure~\ref{fig:hop-transition} illustrates this hop-based reasoning process, highlighting how models transition between documents.

For a question $Q$ and a collection of $m$ documents $D = \{d_1, d_2, \dots, d_m\}$, the task is to predict (1) an answer $A$ (a textual span within one of the documents in $D$), and (2) a reasoning path $\mathcal{P} = (h_1, h_2, \dots, h_{n_{\text{model}}})$ representing the sequence of reasoning hops.

The \textbf{model hop count} is defined as
\begin{equation}
N_{\text{model}} = |\mathcal{P}|,
\end{equation}
where $|\mathcal{P}|$ denotes the length of the model’s hop sequence.

The \textbf{gold hop count} is defined as
\begin{equation}
N_{\text{gold}} = |\mathcal{P}^*|,
\end{equation}
where $\mathcal{P}^*$ is the gold-standard reasoning path required to answer the question.

\subsection{Refining Reasoning Categories}
To diagnose reasoning failures in multi-hop QA, we refined our error taxonomy through three iterative stages. Each stage addressed prior shortcomings and improved inter-annotator agreement, as shown in Figure~\ref{fig:agreement-trend}. Full definitions for Stage 1 and Stage 2 are in the Appendix.

\subsubsection*{Stage 1: Coarse Conceptual Labels}
Our initial taxonomy used four loosely defined labels: \textit{Effective}, \textit{Underthinking}, \textit{Overthinking}, and \textit{Faulty}. These arose from manual trace inspection but lacked clear definitions. Annotators struggled to distinguish between concise reasoning and underthinking, or between verbose, incorrect reasoning and overthinking. The lack of a formal notion of reasoning hops made error tracing difficult. \textit{Faulty} served as a catch-all for various errors, reducing analytical usefulness.

\subsubsection*{Stage 2: Structured Hop-Based Categorization}
In the second stage, we introduced a 10-category taxonomy based on $N_{\text{model}}$, $N_{\text{gold}}$, hop correctness, and answer accuracy to support structured error analysis. As manual evaluation scaled, new ambiguities emerged. Category 8 (early hallucinations) often overlapped with Category 6 (underspecified chains) and question misinterpretation. Annotators also struggled to distinguish shortcut reasoning from flawed logic. These overlaps revealed that even structurally driven categories needed stronger semantic clarity.

\subsubsection*{Stage 3: Final Schema with Meta-Evaluation Markers}
The final schema addressed these issues through clearer definitions. We formally defined a reasoning \textbf{hop}, excluding repeated entity mentions within the same document to avoid inflated hop counts. We also distinguished overthinking via cross-document exploration ($N_{\text{model}}$ > $N_{\text{gold}}$) from verbose or circular reasoning within a document, captured by a separate overthinking flag.

These changes transformed the taxonomy into a more principled and disjoint framework. Annotator agreement improved significantly as category boundaries became more interpretable and semantically grounded.

\subsection{Definitions of Reasoning Categories}
\label{sec:taxonomy}

\begin{table*}[t]
\centering
\renewcommand{\arraystretch}{1.1}
\begin{tabularx}{\textwidth}{p{3.5cm} X}
\toprule
\textbf{Reasoning Category} & \textbf{Definition} \\
\midrule

\parbox[t]{3.5cm}{
$\bm{N}_{\text{model}} = \bm{N}_{\text{gold}};$\\
\textit{Fully Correct Hops}
}
&
The model executes the exact number of required gold reasoning hops, and each hop is logically sound, complete, and correct. \\

\addlinespace

\parbox[t]{3.5cm}{
$\bm{N}_{\text{model}} = \bm{N}_{\text{gold}};$\\
\textit{Partially Correct Hops}
}
&
The model executes the correct number of reasoning steps, but one or more hops involve incorrect documents, entities, or relations. The model reasoning is partially misaligned with the gold reasoning path. \\

\addlinespace

\parbox[t]{3.5cm}{
$\bm{N}_{\text{model}} < \bm{N}_{\text{gold}};$\\
\textit{Fully Correct Hops}
}
&
The model executes fewer hops than required, yet all executed reasoning steps are correct and directly correspond to a subset of the required hops. This indicates incomplete but partially correct reasoning. \\

\addlinespace

\parbox[t]{3.5cm}{
$\bm{N}_{\text{model}} < \bm{N}_{\text{gold}};$\\
\textit{Partially Correct Hops}
}
&
The model executes fewer reasoning steps than required, omitting essential hops and introducing incorrect hops within the shortened chain. The reasoning is both incomplete and partially incorrect. \\

\addlinespace

\parbox[t]{3.5cm}{
$\bm{N}_{\text{model}} > \bm{N}_{\text{gold}};$\\
\textit{Trailing Irrelevance}
}
&
The model initially executes all required reasoning steps but then continues with additional irrelevant hops. These extra steps occur after completing the required reasoning and reflect the model's extraneous elaboration. \\

\addlinespace

\parbox[t]{3.5cm}{
$\bm{N}_{\text{model}} > \bm{N}_{\text{gold}};$\\
\textit{Early Irrelevance}
}
&
The model introduces irrelevant reasoning steps before or interspersed among the required hops. These interruptions disrupt logical reasoning progression, resulting in confusion, distraction or circular reasoning. The required reasoning steps may be partially addressed or incorrect. \\

\addlinespace

\parbox[t]{3.5cm}{
Question\\
Misinterpretation
}
&
The model misunderstands the original question during its early reasoning steps, often focusing on incorrect entities or setting up the wrong task, leading to fundamentally flawed reasoning. \\

\bottomrule
\end{tabularx}
\caption{Definitions of Reasoning Categories in Multi-Hop QA. $N_{\text{model}}$ denotes the number of reasoning hops executed by the model; $N_{\text{gold}}$ is the number of required gold hops.}
\label{tab:reasoning-categories}
\end{table*}

\begin{figure*}[ht]
    \centering
    \includegraphics[width=0.98\textwidth]{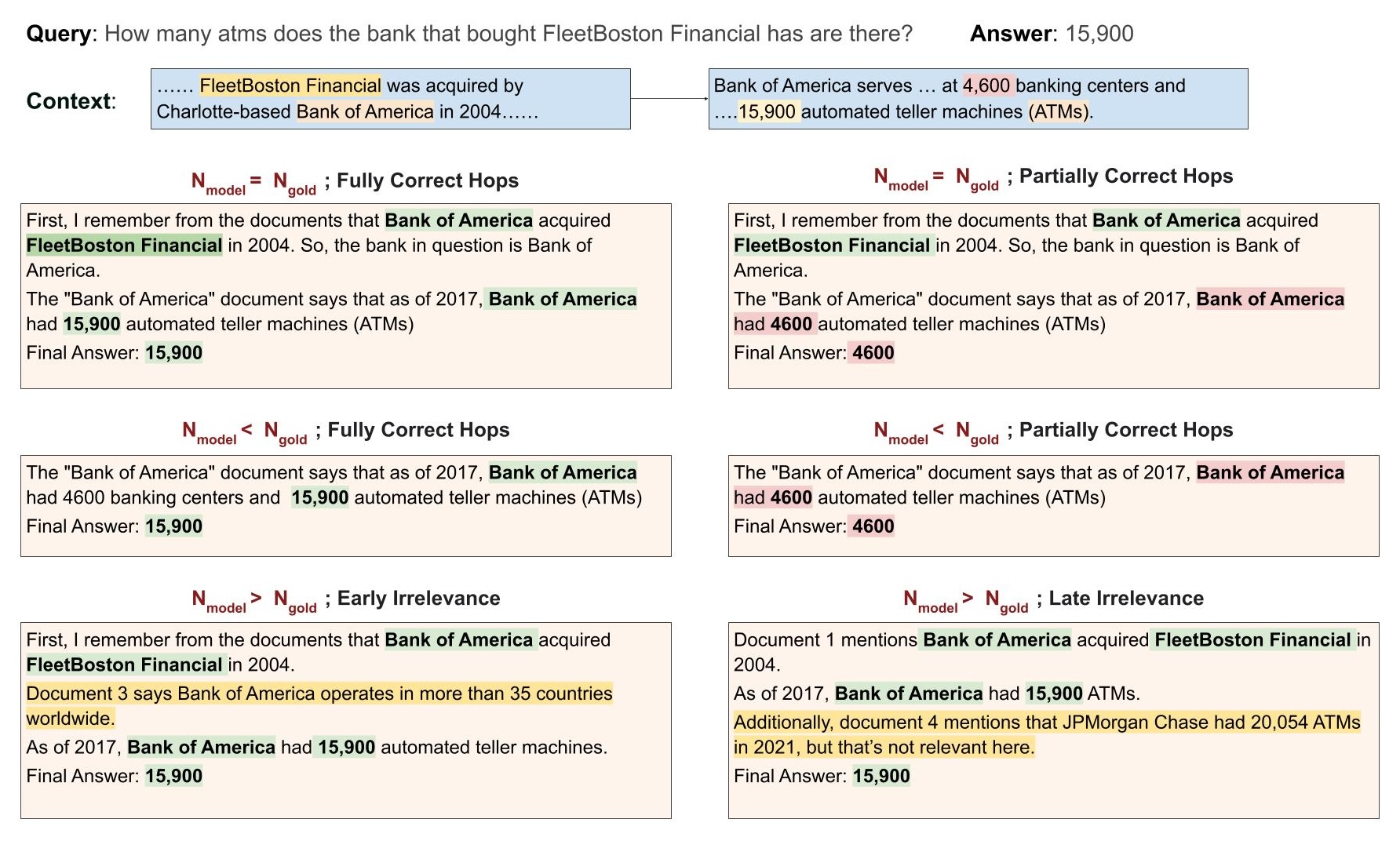} 
    \caption{
    \textbf{Examples of Reasoning Error Categories.}
    Representative outputs illustrating the main error categories in multi-hop reasoning for a single example. The correct entities are highlighted in green, incorrect in red, and irrelevant or extraneous information in yellow.
    }
    \label{fig:error-category-examples}
\end{figure*}

\begin{figure}[ht]
    \centering
    \includegraphics[width=0.4\textwidth]{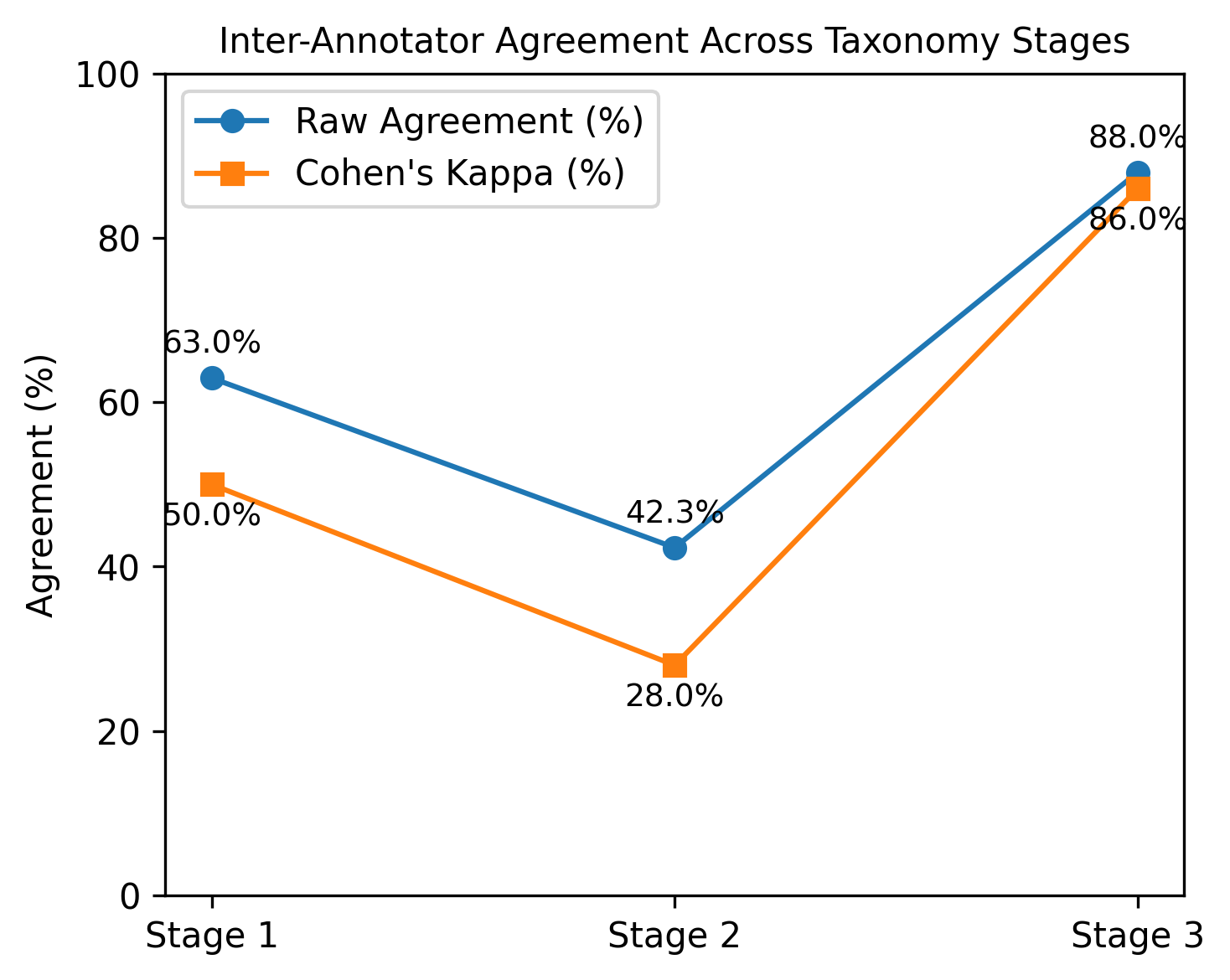}
    \caption{
\textbf{Improvement in Inter-Annotator Agreement Across Refinement Stages.}
Raw agreement and Cohen’s kappa both increase substantially as the reasoning error taxonomy evolves from loosely defined to formally structured categories, with the highest agreement achieved after Stage 3 refinements.
}
    \label{fig:agreement-trend}
\end{figure}

Following iterative refinement and extensive pilot annotations, we arrived at a final taxonomy that enabled high inter-annotator agreement. As shown in Figure~\ref{fig:agreement-trend}, this version resolved prior ambiguities by enforcing stricter hop semantics and introducing meta-evaluation markers to capture surface-level verbosity independently from structural reasoning failure.

Table~\ref{tab:reasoning-categories} summarizes our final taxonomy, providing precise operational definitions for each reasoning error category used in our annotation pipeline. These categories, combined with the meta-evaluation markers of overthinking and coverage, provide comprehensive coverage of potential reasoning errors, enabling systematic and insightful error diagnosis in multi-hop QA models.

\subsubsection*{Meta-Evaluation Markers}
To further enhance our analytical granularity, we introduced meta-evaluation markers:

\paragraph{Overthinking:} This marker captures indicators of cognitive inefficiency in the model's reasoning. It is applied when: 1) the model includes non-essential information from gold documents—such as background details, tangential facts, or calculations—that do not aid in progressing the reasoning chain; and 2) the model demonstrates repetitive or circular behavior, such as repeatedly checking the same entity or relation more than twice.

\paragraph{Coverage:} This marker addresses the completeness of source-document utilization, specifically evaluating whether the model successfully retrieves all necessary source documents. Low coverage indicates gaps in retrieval or attention, leading to incomplete reasoning chains or unsupported conclusions.

\section{Experimental Setup}

\paragraph{Models}
We analyze six language models that span a range of architectures, parameter scales, and accessibility. Our primary focus is on four open-source distilled models—\textsc{DeepSeek-R1-Distill-LLaMA-8B}, \textsc{DeepSeek-R1-Distill-LLaMA-70B}, \textsc{DeepSeek-R1-Distill-Qwen-7B}, and \textsc{DeepSeek-R1-Distill-Qwen-14B}. To complement these, we include two original reasoning models: \textsc{Claude 3.7 Sonnet}, a proprietary reasoning model, and \textsc{DeepSeek-R1}, an open-weight reasoning model.

For all DeepSeek models, we set the generation temperature to 0.6, following the recommendations of \citet{liu2024deepseek}, to mitigate endless repetition or incoherent outputs. For Claude 3.7 Sonnet, we use a deterministic setting with the temperature set to 0.

\paragraph{Datasets}
We evaluate model reasoning across three multi-hop QA datasets of increasing difficulty: 2WikiMultiHopQA~\cite{ho2020constructingmultihopqadataset}, which emphasizes structured multi-hop reasoning; HotpotQA~\cite{hotpotqa}, which includes distractors and diverse reasoning types like comparisons; and MuSiQue~\cite{musique}, a high-complexity benchmark designed to minimize shortcuts through dense context and sub-question dependencies. Dataset details are provided in Table~\ref{tab:dataset-stats} in the appendix.


\subsubsection*{Question Types}

To enable systematic reasoning analysis, we categorize multi-hop questions into five distinct types based on their logical structure: \textit{Compositional}, \textit{Comparison}, \textit{Intersection}, \textit{Inference}, and \textit{Bridge Comparison}. These categories reflect the types of reasoning steps required to arrive at the correct answer.

Detailed definitions and illustrative examples for each type are provided in Appendix (Table~\ref{Appendix:question-types}).

\subsection{Annotation Process}
\label{sec:annotation}

To evaluate the reasoning quality of model outputs, we develop a structured human annotation pipeline encompassing three key stages:

    \paragraph{Sampling and Generation:}  
    We uniformly sampled 240 questions across HotpotQA, 2Wiki-MultiHopQA, and MuSiQue. Six models answered each question using a standardized prompting strategy designed to minimize instruction-induced bias. 

    \paragraph{Final Answer and Meta Eval Markers:}  
    The final answers were evaluated for correctness using automated matching, with manual verification for paraphrased or non-exact responses. Simultaneously, we annotated: 1) the \(N_{\text{model}}\),  
    2) the binary \textit{Coverage} Marker and 3) the \textit{Overthinking Marker} .  

    \paragraph{Reasoning Category Assignment:}  
    Each response was categorized into one of our predefined reasoning error types. (see Section~\ref{sec:taxonomy}).

Figures~\ref{fig:interface1} and \ref{fig:interface2} show our custom annotation interface, which facilitates structured error labeling, flag toggling, and hop trace visualization.
In total, we annotated 1,440 model outputs. Following the removal of examples with missing answers in the context caused by dataset artifacts, 1,080 examples were retained for analysis.

\section{Human Evaluation Results}





\subsection{Reasoning Fidelity and Answer Accuracy}

\begin{figure*}[htbp]
    \centering
    \begin{minipage}[b]{0.32\textwidth}
        \centering
        \includegraphics[width=\linewidth]{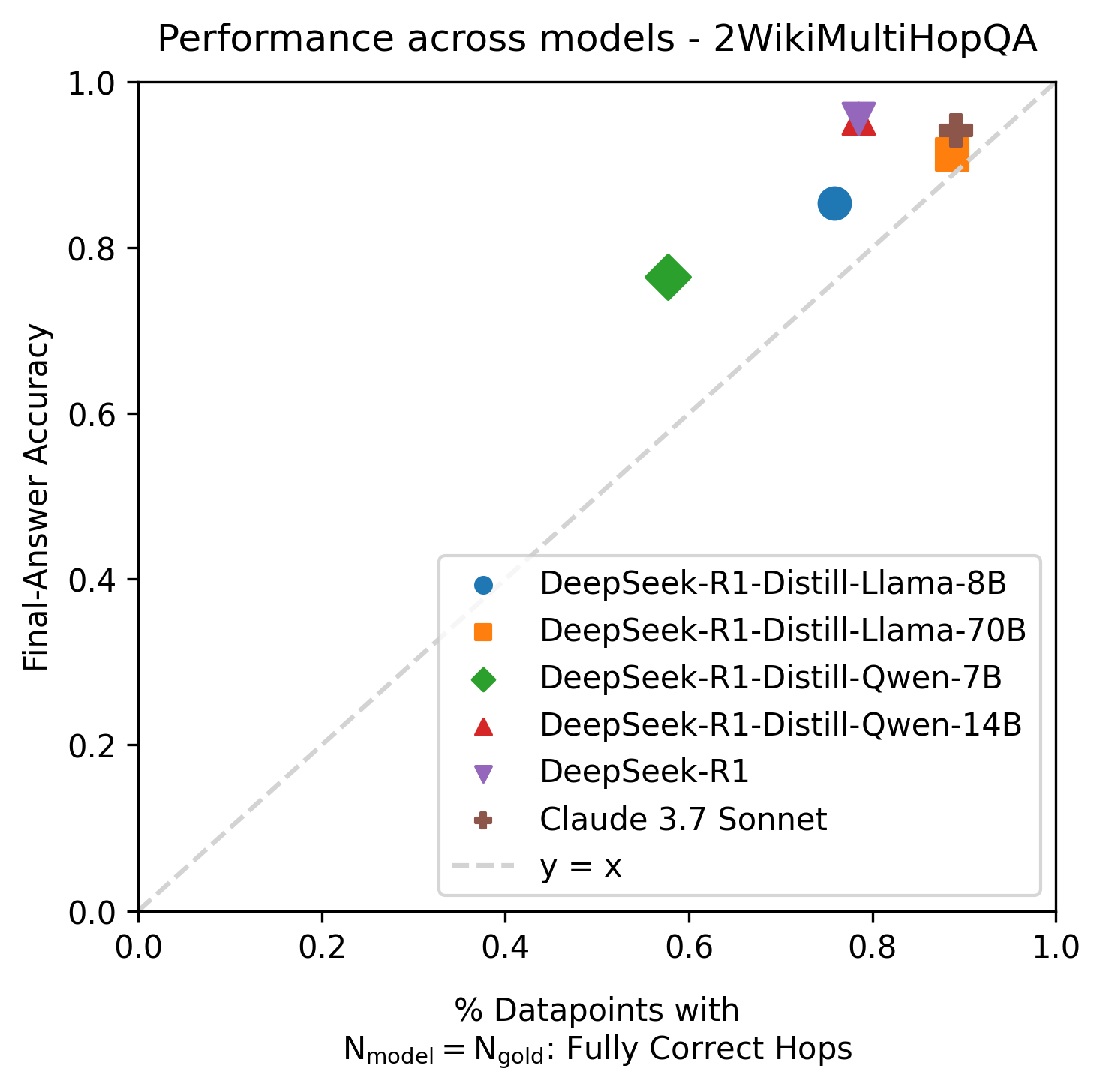}
        \caption*{(a) 2Wiki}
    \end{minipage}
    \hfill
    \begin{minipage}[b]{0.32\textwidth}
        \centering
        \includegraphics[width=\linewidth]{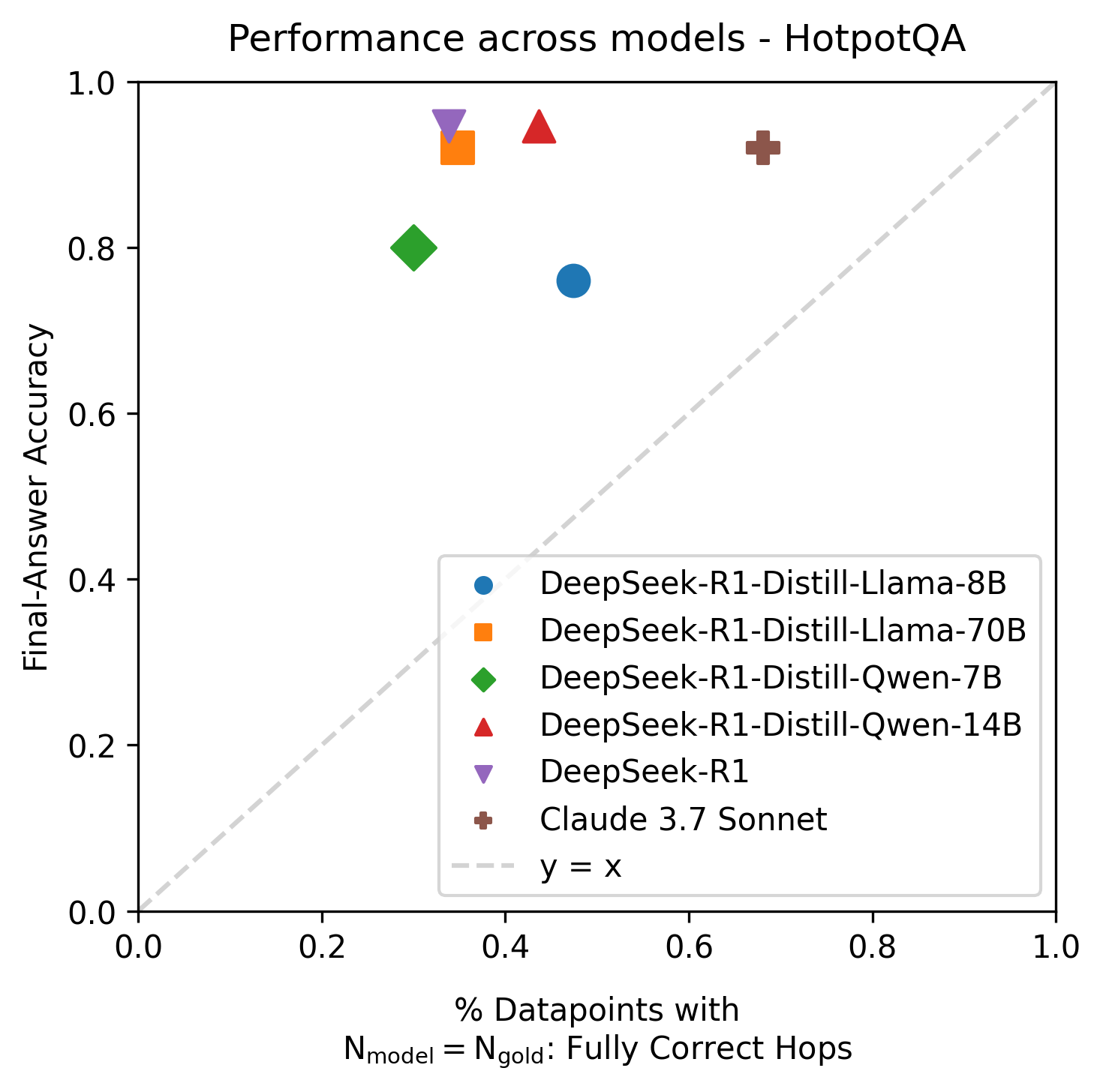}
        \caption*{(b) HotpotQA}
    \end{minipage}
    \hfill
    \begin{minipage}[b]{0.32\textwidth}
        \centering
        \includegraphics[width=\linewidth]{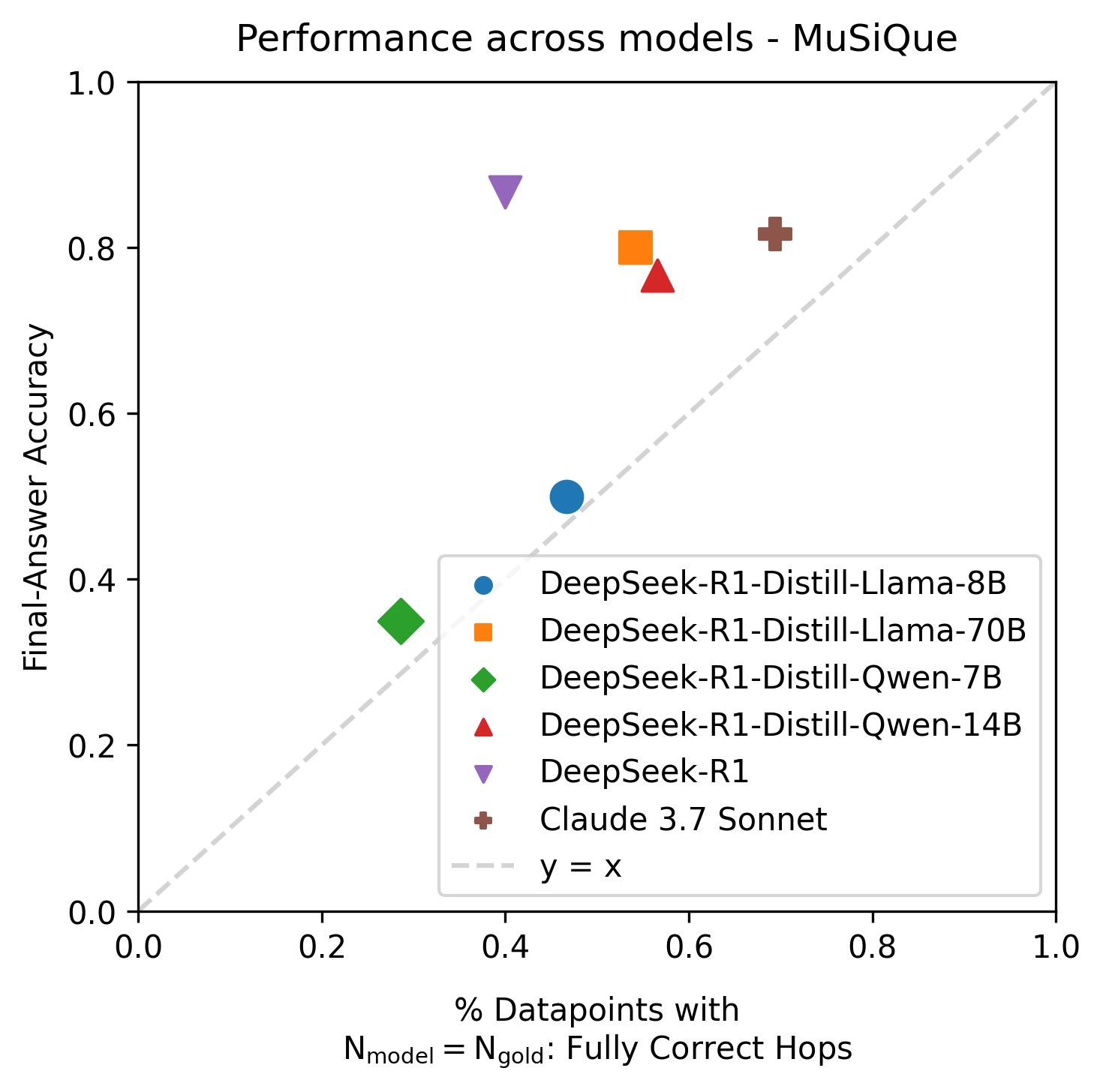}
        \caption*{(c) MuSiQue}
    \end{minipage}
    \caption{
\textbf{Relationship Between Reasoning Fidelity and Answer Accuracy Across Datasets.}
Each subplot shows model performance on (a) 2Wiki, (b) HotpotQA, and (c) MuSiQue. Each point represents the performance of a model, with the x-axis showing the fraction of fully correct reasoning traces (\(N_{\text{model}}=N_{\text{gold}}\)) and the y-axis showing final answer accuracy. The dotted diagonal (\(y = x\)) marks perfect alignment; points above the line indicate models that answer correctly even when reasoning is imperfect.
}
    \label{fig:performance-trio}
\end{figure*}

\noindent Figure~\ref{fig:performance-trio} summarizes model behavior on fully correct hop alignment (\(N_{\text{model}}=N_{\text{gold}}\)) and final-answer accuracy across datasets.  We see that reasoning Fidelity holds in Simpler Tasks but collapses in Complex Chains. Across all datasets, Claude 3.7 achieves the highest accuracy.


\paragraph{High Reasoning Fidelity on 2Wiki:} All models perform strongly on the 2Wiki dataset, with the majority of models have around 80\% datapoints with \( N_{\text{model}} = N_{\text{gold}} \)correspondence: fully correct hops, and near-perfect final answer accuracy  This confirms that current LMs reliably handle simple multi-hop questions.

\paragraph{Inefficient Reasoning in HotpotQA: } Performance on HotpotQA shows the highest concentration of (\(N_{\text{model}}>N_{\text{gold}}\))
 (Figure~\ref{fig:reasoning-errors-hotpot}). While the final-answer accuracy remains high, the presence of semantically dense and distractor-filled paragraphs leads models to over-explore the context, often beyond the required inference chain. This behavior highlights the limitations of current LMs in maintaining focused reasoning under noisy, multi-document settings.

 \paragraph{Intermediate Fidelity and Model-Specific Patterns emerge on the MuSiQue dataset: } Larger models demonstrate intermediate reasoning fidelity (45–65\%) alongside relatively high answer accuracy. Smaller models exhibit poor performance on both metrics, underscoring difficulties in complex multi-hop contexts. Notably, DeepSeek-R1 shows the greatest divergence, achieving very high answer accuracy despite substantially lower reasoning fidelity.


\subsection{Reasoning Patterns Across Models and Datasets}

Figure~\ref{fig:reasoning-errors-all} show the distribution of reasoning error types in the MuSiQue, 2Wiki-MultiHopQA, and HotpotQA datasets. Our analysis reveals the following insights:

\paragraph{Claude 3.7 Sonnet Sets the Bar for Stable and Precise Reasoning: }
Among all evaluated models, Claude 3.7 Sonnet demonstrates the most stable and controlled reasoning behavior. It consistently maintains high rates of fully correct reasoning while keeping all other error types—especially early and trailing irrelevance—significantly lower than both DeepSeek-R1 and the distilled model variants. 


\paragraph{Overhopping is the Most Persistent and Systemic Reasoning Failure: }
Across all datasets and models, overhopping (\( N_{\text{model}} > N_{\text{gold}} \) categories in Figure ~\ref{fig:reasoning-errors-all}) is consistently higher than other errors. This often stems from contextual redundancy or ambiguity, pushing models to over-explore rather than terminate. The Qwen family of models particularly struggles with this issue: frequently displaying early and trailing irrelevance errors—even at larger scales—indicating a proclivity for recall over precise reasoning.



\paragraph{Scaling Models Improves Simple Reasoning but Leaves Complex Errors Unresolved:}
As shown in Figure~\ref{fig:reasoning-errors-all}, increasing model size leads to more examples with fully correct hops (the leftmost bars in each subplot), particularly on simpler tasks like 2Wiki (Figure~\ref{fig:reasoning-errors-all}a). However, for more complex datasets such as HotpotQA and MuSiQue (Figure~\ref{fig:reasoning-errors-all}b,~c), the gains from scaling plateau. Even the largest models still exhibit substantial numbers of early irrelevance and trailing irrelevance errors (the right-side bars). This persistent error pattern indicates that, while scale enhances basic multi-hop reasoning, it does not fully resolve deeper reasoning challenges in complex or distractor-heavy settings.

\paragraph{Deepseek-R1 Distilled Models Rival the Deepseek-R1 Counterpart in Multi-hop Tasks: }
On both simple and moderately complex datasets, distilled LLaMA variants show strong reasoning alignment. The LLaMA 70B variant performs almost similarly or even better than the original Deepseek-R1 model.



\subsection{Relationship Between Reasoning Errors and Final Answer Correctness}

Figure~\ref{fig:answer-correctness-all} examines the relationship between reasoning trace quality and final answer correctness across datasets.

\paragraph{Correctness Tightly Coupled with Reasoning Quality:}
As seen in the leftmost bars in Figure~\ref{fig:answer-correctness-all}, across all datasets, final correct answers almost exclusively emerge from the (\( N_{\text{model}} = N_{\text{gold}} \)), Fully Correct Hops category. Even minor deviations in the reasoning chain, such as exceeding hops or partial hops, reduce the likelihood of correctness.

\paragraph{Answer Correctness is Sensitive to Missing Hops:}
Looking at the "Partially Correct Hops" bars in all three panels of Figure~\ref{fig:answer-correctness-all}, we see that incomplete reasoning rarely yields a correct answer. This confirms that failure to cover all necessary facts, even in part, is a definitive bottleneck in LMs’ reasoning chains.

\paragraph{Smaller Models are More Fragile to Reasoning Errors:}
As shown in Figure~\ref{fig:answer-correctness-all} across all datasets, smaller models such as LLaMA-8B and Qwen-7B exhibit a higher propensity for reasoning errors to cascade into incorrect final answers. In contrast, larger models like DeepSeek-R1 and Claude 3 Sonnet demonstrate greater robustness, with fewer incorrect answers arising from these types of reasoning errors.

\paragraph{Early Irrelevance is More Detrimental than Trailing Irrelevance:}
In every panel of Figure~\ref{fig:answer-correctness-all}, the "Early Irrelevance" category shows that "Answer Incorrect" bars are higher compared to that corresponding to "Trailing Irrelevance." This suggests that irrelevant reasoning steps introduced early in the chain are more disruptive
to the model’s final answer.

\begin{table*}[ht]
    \centering
    \begin{tabular}{lccc}
        \toprule
        \textbf{Model} & \textbf{2Wiki-MultiHopQA} & \textbf{HotpotQA} & \textbf{MuSiQue} \\
        \midrule
        DeepSeek-R1-Distill-Llama-8B      & 41.2\% & 29.3\% & 48.3\% \\
        DeepSeek-R1-Distill-Llama-70B     & 19.1\% & 12.0\% & 41.7\% \\
        DeepSeek-R1-Distill-Qwen-7B       & 26.5\% & 41.3\% & 61.7\% \\
        DeepSeek-R1-Distill-Qwen-14B      & 30.9\% & 28.0\% & 50.0\% \\
        DeepSeek-R1                       & 27.9\% & 18.7\% & 53.3\% \\
        Claude 3.7 Sonnet                 & 22.1\% & 22.7\% & 36.7\% \\
        \bottomrule
    \end{tabular}
    \caption{
\textbf{Overthinking Rates by Model and Dataset.}
Percentage of answers with Overthinking for each model on the 2Wiki-MultiHopQA, HotpotQA, and MuSiQue datasets. The results highlight the substantial increase in overthinking in more complex MuSiQue dataset.
}
\label{tab:overthinking-percentage}
\end{table*}

\subsection{Overthinking Trends and Their Impact}

We systematically examine the prevalence and impact of overthinking across different models and datasets, highlighting how this phenomenon influences overall model performance and error rates.

 \paragraph{Overthinking Surges in Complex Reasoning Tasks:} 
As shown in the MuSiQue results (see Figure~\ref{fig:reasoning-errors-all}c and Table~\ref{tab:overthinking-percentage}), overthinking rises markedly across all models, with rates ranging from 36.7\% to 61.7\%. Notably, DeepSeek-R1-Distill-Qwen-7B reaches the highest overthinking rate of 61.7\%, while even advanced models such as Claude 3 Sonnet and DeepSeek-R1 exhibit elevated rates. This trend suggests that task complexity, rather than model scale, is the primary driver of overthinking.

\paragraph{Overthinking is a Systematic Source of Incorrect Answers:} 
A significant portion of incorrect answers are accompanied by overthinking, especially in MuSiQue (see Figure~\ref{fig:answer-incorrectness-musique-overthink}). Although HotpotQA and 2Wiki contain fewer errors labeled as overthinking (see Table~\ref{tab:overthinking-percentage}), when overthinking does occur, it almost always results in incorrect answers (see shaded bars in Figure~\ref{fig:answer-incorrectness-2wiki-overthink} and Figure~\ref{fig:answer-incorrectness-hotpot-overthink}). This finding suggests that the negative impact of overthinking is not just limited to complex datasets but also arises from the logical incoherence it introduces, irrespective of task difficulty. Overthinking is not merely harmless elaboration, but a systematic driver of reasoning collapse and failure to reach a final answer.

\begin{figure*}[htbp]
    \centering
    \includegraphics[width=\linewidth]{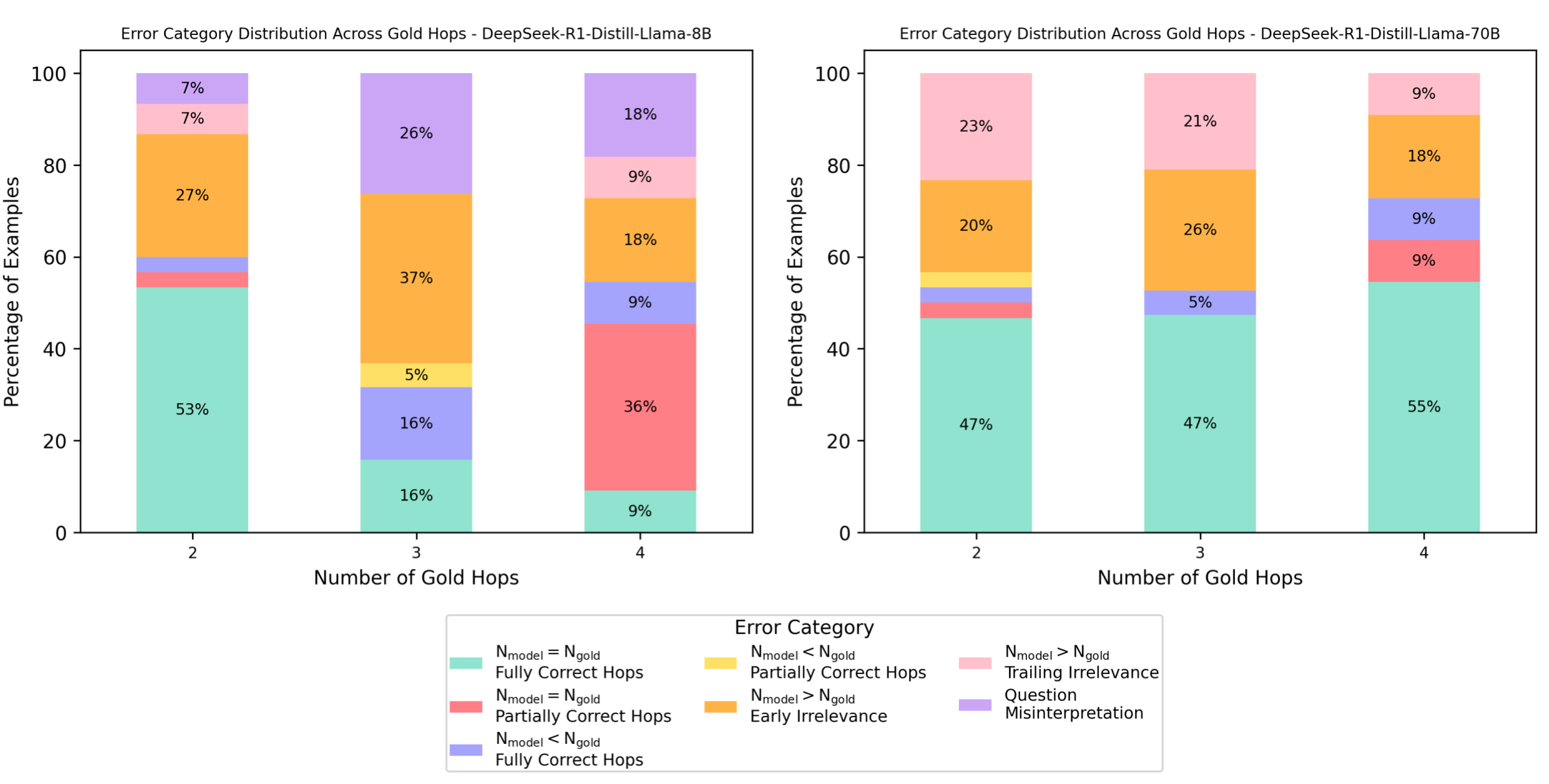}
    \caption{
\textbf{Hop-wise Distribution of Reasoning Errors in LLaMA Family Models.}
The left and right panels compare error trends for LLaMA-8B and LLaMA-70B, respectively. Results highlight the decline in fully correct reasoning with greater hop count, and the increasing prevalence of overhopping errors on harder questions.
}
\label{fig:hopwise-duo}
\end{figure*}

\subsection{Distribution across Question types}
Across the datasets, the multi-hop questions can be grouped into four key categories: \textit{Bridge Comparison questions}, \textit{Comparison questions}, \textit{Compositional questions}, and \textit{Inference Questions} (details of the question types and examples are in Section \ref{Appendix:question-types}).

Figure~\ref{fig:qtypes_error_all} shows the distribution of reasoning error types across question categories for all the models. We observe the following trends across different question types:

\paragraph{Bridge Comparison Questions Are Consistently Solved, Especially from 2Wiki:} Bridge questions (mainly from 2Wiki) yield 94–100\% fully correct hops across all models. Even smaller models like Qwen-7B and LLaMA-8B perform well, while Claude 3.7 Sonnet and Qwen-14B make no errors.  These questions often contain explicit reference to entities or co-occurrence patterns that mirror the pre-training distribution of the model, allowing models to resolve them through recognition of patterns at the surface level rather than deep reasoning.
    
\paragraph{Symmetric Structures Trigger Redundant Reasoning and Overhopping:} Found in HotpotQA and 2Wiki, comparison questions show 50–68\% fully correct rates, with 25–45\% of errors due to early or trailing irrelevance. Their symmetric phrasing encourages exploration of both options, even when one suffices. Claude occasionally bypasses intermediate hops while still producing correct answers, suggesting reliance on shortcut-style or selective reasoning paths.

\paragraph{Compositional Reasoning Exposes Integration Failures:} Compositional questions strain models’ ability to synthesize disjoint facts. Smaller models (Qwen-7B, LLaMA-8B, DeepSeek-R1) show many partially correct chains, even with correct hop counts. Claude and LLaMA-70B perform better, suggesting that scale and architecture improve integration. Misinterpretation errors are also higher here.

\paragraph{Inference Questions Are the Most Error-Prone and Trigger Overthinking:} Inference questions, heavily present in MuSiQue and 2Wiki, demand implicit reasoning and multi-step logic without strong lexical cues. These questions yield the broadest error types, early/trailing irrelevance, misinterpretation, and underhopping. Qwen-7B answers only 10\% correctly, with 30\% misinterpretation. Even DeepSeek-R1 shows 37\% trailing irrelevance. Only Claude and LLaMA-70B manage modest control (50–55\% correct), highlighting the inherent difficulty of inference.

\paragraph{Inference and Compositional Tasks Drive Overthinking:} Overhopping is most common in inference questions, reaching 70\% in Qwen-7B, 65\% in LLaMA-8B, and ~60\% in DeepSeek-R1 and Qwen-14B. Lack of clear stopping cues leads models to overgenerate. Bridge questions show minimal overhopping (<20\%) due to their bounded structure.

\paragraph{Even Large Models Struggle to Combine Retrieved Evidence:} In compositional and inference settings, all models — including large ones like DeepSeek-R1 and Claude 3.7 Sonnet — exhibit partially correct reasoning chains, even when hop counts match the gold labels. This reveals that accurate retrieval alone is insufficient; models must also effectively link and reason over retrieved evidence. These synthesis failures suggest weaknesses in chain-of-thought alignment and reasoning coordination, particularly under pressure from semantically distant facts.




\subsection{Hop-wise Error Distribution}

To better understand how reasoning evolves across multi-hop inference chains, we analyze the distribution of reasoning errors at the hop level. Figure~\ref{fig:hopwise-duo} presents hop-wise error trends for the LLaMA family, while Figure~\ref{fig:hopwise_all} illustrates these trends for other models.

\paragraph{Larger Models Are More Stable Across Hop Counts:}
As the number of required reasoning steps increases, most models exhibit a clear drop in fully correct reasoning (\(N_{\text{model}} = N_{\text{gold}}\)). For example, in Figure~\ref{fig:hopwise-duo} (left panel), DeepSeek-R1-Distill-Llama-8B achieves 53\% accuracy on 2-hop questions, but this drops to 16\% for 3-hop and just 9\% for 4-hop examples. In contrast, larger models such as DeepSeek-R1-Distill-Llama-70B (Figure~\ref{fig:hopwise-duo}, right panel) and Claude 3.7 Sonnet (Figure~\ref{fig:hopwise_all}d) show much greater stability across hop lengths, maintaining relatively consistent performance even as reasoning depth increases. This suggests these models have a stronger capacity to follow and complete longer reasoning chains without deviation.

\paragraph{Overhopping Is a Major Error Source in Harder Questions:}
For 4-hop questions, the most prominent error across several models is early irrelevance (\(N_{\text{model}} > N_{\text{gold}}\)). This is especially clear for DeepSeek-R1-Distill-Qwen-7B (Figure~\ref{fig:hopwise_all}a), where 73\% of 4-hop examples are categorized as early irrelevance. Both Claude 3.7 Sonnet and DeepSeek-R1-Distill-Qwen-14B (Figure~\ref{fig:hopwise_all}b and d) show 45\% early irrelevance at 4 hops. These results indicate that, in more complex tasks, models frequently continue reasoning beyond what is necessary, retrieving irrelevant or redundant information.

\paragraph{Shallow Collapse in Qwen-7B, Depth Limitations in Claude 3.7 Sonnet: }
Qwen-7B (Figure~\ref{fig:hopwise_all}a) shows signs of partial reasoning at 3 hops but collapses almost entirely into early irrelevance (73\%) at 4 hops, abandoning intermediate reasoning strategies. This suggests that, under high reasoning load, smaller models tend to default to over-retrieval. In contrast, Claude 3.7 Sonnet (Figure~\ref{fig:hopwise_all}d) maintains strong performance up to 3 hops but shows a spike in early irrelevance (45\%) at 4 hops. Even advanced models, therefore, encounter depth calibration issues, struggling to determine when to stop in extended reasoning chains.

\section{Automated Evaluation Results}

\begin{figure}[htbp]
    \centering
    \includegraphics[width=0.9\linewidth]{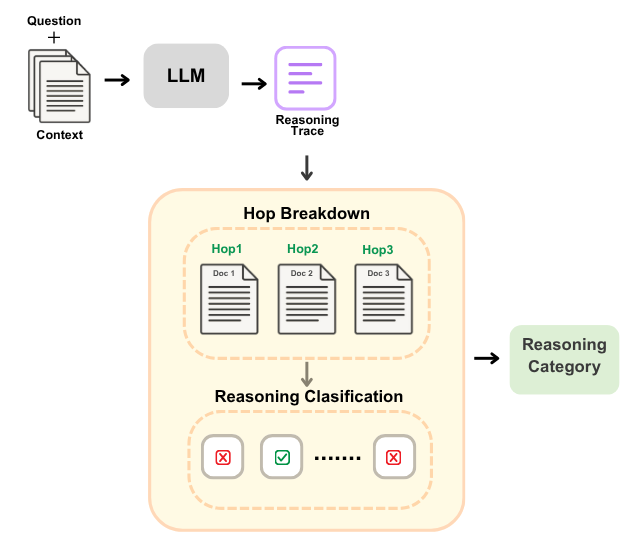}
    \caption{
\textbf{Two-Step LLM-Assisted Evaluation Workflow.}
A high-level overview of the two-step decomposition that improves annotation accuracy and consistency for complex multi-hop reasoning tasks.
}
    \label{fig:autoeval}
\end{figure}

While extensive manual evaluations provide detailed and reliable insights into model reasoning behaviors, it is difficult to scale, particularly for complex queries from datasets like MuSiQue, where each annotation can take approximately four minutes per data point. Hence, We develop a framework for automating the annotation process to significantly improve evaluation efficiency.

\subsection{Evaluation Workflow}

We employed an LLM-as-a-Judge framework to automate the annotation task, using \texttt{gpt-4.1-mini}\footnote{
A state-of-the-art model that is different from the models analyzed in this work.}
as our judging model. Utilizing LLM as a judge for annotating reasoning failures helped us scale the process significantly and reduced evaluation time, achieving approximately a \textbf{20x} increase in efficiency compared to manual annotation. To ensure parity with manual annotation process and high fidelity  analysis, we provided the Judge LLM with the same detailed annotation guidelines used by human annotators, but prompt-engineered the guidelines with explicit formatting instructions and clear definitions. The Judge LLM had access to 'question', 'relevant context documents', and the 'final response' from the reasoning models.



Consistent with findings for multi-step judging process in \cite{wang2023selfconsistencyimproveschainthought, zheng2024processbenchidentifyingprocesserrors}, we adopt a two-step annotation process as illustrated in Figure \ref{fig:autoeval}:

\begin{enumerate}

\item \textbf{Hop Breakdown}: First, the Judge LLM identifies and annotates the reasoning hops present in the model's response.

\item \textbf{Reasoning Classification}: Next, the Judge uses these annotated hops to accurately categorize the response into one of our predefined error categories.

\end{enumerate}
This decomposition significantly improved annotation accuracy and consistency, aligning with findings from recent literature indicating that multi-step judging processes enhance reliability and accuracy in complex evaluation tasks \cite{wang2022self}.


\subsection{Model-Wise Agreement with Human Annotations}

To further validate our LLM-as-a-Judge pipeline, we evaluate the consistency of annotations across six models and three datasets: \textsc{MuSiQue}, \textsc{2Wiki}, and \textsc{HotpotQA}. Based on the results presented in Table~\ref{tab:judge_model_agreement}, our LLM-as-a-Judge framework demonstrates promising potential for automating error categorization tasks traditionally performed by human annotators. Across all models and datasets, agreement rates vary, indicating model-specific and dataset-specific challenges. For example, models like DeepSeek-R1 and LLaMA 70B exhibit notably higher agreement rates, particularly on simpler datasets like \textsc{2Wiki}, achieving above 90\%. Conversely, the more challenging \textsc{MuSiQue} dataset consistently shows lower agreement scores, underscoring inherent complexities and subtle reasoning errors that the Judge LLM struggles to replicate accurately.

These results imply that while LLM-as-a-Judge systems are highly effective at automating error categorization for straightforward multi-hop reasoning tasks, complexities in certain datasets highlight the continuing necessity for human judgment or advanced refinement of Judge LLM instructions. The observed variability underscores that further investigation is essential to understand and mitigate factors contributing to lower Judge-model agreement rates, such as nuanced reasoning steps or subtle misinterpretations. Nonetheless, the substantial reduction in annotation time and generally high fidelity in simpler contexts strongly support the viability and efficiency of integrating LLM-as-a-Judge frameworks into broader NLP evaluation pipelines.

\begin{table}[h]
\centering
\caption{LLM-as-a-Judge Agreement (\%) with Human Annotations Across Models and Datasets}
\label{tab:judge_model_agreement}
\begin{tabular}{lrrr}
\toprule
      \textbf{Model} & \textbf{HotpotQA} & \textbf{2Wiki} & \textbf{MuSiQue} \\
\midrule
   LLaMA 8B    & 65.3 & 73.5 & 53.3 \\
   DeepSeek-R1 & 76.0 & 91.1 & 62.6 \\
   LLaMA 70B   & 72.0 & 92.6 & 75.0 \\
   Qwen 7B     & 66.6 & 75.0 & 46.6 \\
   Claude 3.7  & 73.3 & 91.1 & 76.6 \\
   Qwen 14B    & 65.3 & 88.2 & 78.3 \\
\bottomrule
\end{tabular}
\end{table}



\section{Conclusion}
We introduce a hop-based diagnostic framework for multi-hop QA that captures reasoning fidelity through fine-grained error categories and meta-markers for coverage and overthinking. Analysis of six LMs across three datasets reveals high fidelity in simple settings but persistent overhopping, misinterpretation, and synthesis failures in complex and distractor-rich tasks. Our two-step LLM-as-a-Judge method achieves up to 92\% agreement with humans on simpler datasets while cutting evaluation time by 20x, although challenges remain for nuanced reasoning. These findings call for evaluation and training strategies that bridge the gap between correct answers and reasoning that is both efficient and faithful for truly reliable multi-hop QA systems.
\bibliography{custom}

\begin{thebibliography}{21}
\providecommand{\natexlab}[1]{#1}

\bibitem[{Agarwal et~al.(2024)Agarwal, Tanneru, and Lakkaraju}]{faithfulnessvsplausibilityunreliability}
Chirag Agarwal, Sree~Harsha Tanneru, and Himabindu Lakkaraju. 2024.
\newblock \href {https://arxiv.org/abs/2402.04614} {Faithfulness vs. plausibility: On the (un)reliability of explanations from large language models}.
\newblock \emph{Preprint}, arXiv:2402.04614.

\bibitem[{Bao et~al.(2024)Bao, Zhang, Wang, Yang, and Zhang}]{bao2024likelyllmscotmimic}
Guangsheng Bao, Hongbo Zhang, Cunxiang Wang, Linyi Yang, and Yue Zhang. 2024.
\newblock \href {https://arxiv.org/abs/2402.16048} {How likely do llms with cot mimic human reasoning?}
\newblock \emph{Preprint}, arXiv:2402.16048.

\bibitem[{Ho et~al.(2020)Ho, Nguyen, Sugawara, and Aizawa}]{ho2020constructingmultihopqadataset}
Xanh Ho, Anh-Khoa~Duong Nguyen, Saku Sugawara, and Akiko Aizawa. 2020.
\newblock \href {https://arxiv.org/abs/2011.01060} {Constructing a multi-hop qa dataset for comprehensive evaluation of reasoning steps}.
\newblock \emph{Preprint}, arXiv:2011.01060.

\bibitem[{Ishii et~al.(2024)Ishii, Inoue, Suzuki, and Sekine}]{ishii-etal-2024-analysis}
Ai~Ishii, Naoya Inoue, Hisami Suzuki, and Satoshi Sekine. 2024.
\newblock \href {https://doi.org/10.18653/v1/2024.kallm-1.3} {Analysis of {LLM}`s {\textquotedblleft}spurious{\textquotedblright} correct answers using evidence information of multi-hop {QA} datasets}.
\newblock In \emph{Proceedings of the 1st Workshop on Knowledge Graphs and Large Language Models (KaLLM 2024)}, pages 24--34, Bangkok, Thailand. Association for Computational Linguistics.

\bibitem[{Kamoi et~al.(2024)Kamoi, Das, Lou, Ahn, Zhao, Lu, Zhang, Zhang, Zhang, Vummanthala, Dave, Qin, Cohan, Yin, and Zhang}]{kamoi2024evaluatingllmsdetectingerrors}
Ryo Kamoi, Sarkar Snigdha~Sarathi Das, Renze Lou, Jihyun~Janice Ahn, Yilun Zhao, Xiaoxin Lu, Nan Zhang, Yusen Zhang, Ranran~Haoran Zhang, Sujeeth~Reddy Vummanthala, Salika Dave, Shaobo Qin, Arman Cohan, Wenpeng Yin, and Rui Zhang. 2024.
\newblock \href {https://arxiv.org/abs/2404.03602} {Evaluating llms at detecting errors in llm responses}.
\newblock \emph{Preprint}, arXiv:2404.03602.

\bibitem[{Lanham et~al.(2023)Lanham, Chen, Radhakrishnan, Steiner, Denison, Hernandez, Li, Durmus, Hubinger, Kernion, Lukošiūtė, Nguyen, Cheng, Joseph, Schiefer, Rausch, Larson, McCandlish, Kundu, Kadavath, Yang, Henighan, Maxwell, Telleen-Lawton, Hume, Hatfield-Dodds, Kaplan, Brauner, Bowman, and Perez}]{lanham2023measuringfaithfulnesschainofthoughtreasoning}
Tamera Lanham, Anna Chen, Ansh Radhakrishnan, Benoit Steiner, Carson Denison, Danny Hernandez, Dustin Li, Esin Durmus, Evan Hubinger, Jackson Kernion, Kamilė Lukošiūtė, Karina Nguyen, Newton Cheng, Nicholas Joseph, Nicholas Schiefer, Oliver Rausch, Robin Larson, Sam McCandlish, Sandipan Kundu, and 11 others. 2023.
\newblock \href {https://arxiv.org/abs/2307.13702} {Measuring faithfulness in chain-of-thought reasoning}.
\newblock \emph{Preprint}, arXiv:2307.13702.

\bibitem[{Li et~al.(2024)Li, Wang, Li, Guo, Zhang, and Feng}]{li2024evaluatingmathematicalreasoninglarge}
Xiaoyuan Li, Wenjie Wang, Moxin Li, Junrong Guo, Yang Zhang, and Fuli Feng. 2024.
\newblock \href {https://arxiv.org/abs/2406.00755} {Evaluating mathematical reasoning of large language models: A focus on error identification and correction}.
\newblock \emph{Preprint}, arXiv:2406.00755.

\bibitem[{Liu et~al.(2024)Liu, Feng, Xue, Wang, Wu, Lu, Zhao, Deng, Zhang, Ruan et~al.}]{liu2024deepseek}
Aixin Liu, Bei Feng, Bing Xue, Bingxuan Wang, Bochao Wu, Chengda Lu, Chenggang Zhao, Chengqi Deng, Chenyu Zhang, Chong Ruan, and 1 others. 2024.
\newblock Deepseek-v3 technical report.
\newblock \emph{arXiv preprint arXiv:2412.19437}.

\bibitem[{Mahdavi et~al.(2025)Mahdavi, Hashemi, Daliri, Mohammadipour, Farhadi, Malek, Yazdanifard, Khasahmadi, and Honavar}]{mahdavi2025brainsvsbytesevaluating}
Hamed Mahdavi, Alireza Hashemi, Majid Daliri, Pegah Mohammadipour, Alireza Farhadi, Samira Malek, Yekta Yazdanifard, Amir Khasahmadi, and Vasant Honavar. 2025.
\newblock \href {https://arxiv.org/abs/2504.01995} {Brains vs. bytes: Evaluating llm proficiency in olympiad mathematics}.
\newblock \emph{Preprint}, arXiv:2504.01995.

\bibitem[{Mukherjee et~al.(2025)Mukherjee, Chinta, Kim, Sharma, and Hakkani-Tür}]{mukherjee2025premiseaugmentedreasoningchainsimprove}
Sagnik Mukherjee, Abhinav Chinta, Takyoung Kim, Tarun~Anoop Sharma, and Dilek Hakkani-Tür. 2025.
\newblock \href {https://arxiv.org/abs/2502.02362} {Premise-augmented reasoning chains improve error identification in math reasoning with llms}.
\newblock \emph{Preprint}, arXiv:2502.02362.

\bibitem[{Sakarvadia(2024)}]{interpretinglanguagemodelscase}
Mansi Sakarvadia. 2024.
\newblock \href {https://arxiv.org/abs/2411.05037} {Towards interpreting language models: A case study in multi-hop reasoning}.
\newblock \emph{Preprint}, arXiv:2411.05037.

\bibitem[{Shojaee et~al.(2025)Shojaee, Mirzadeh, Alizadeh, Horton, Bengio, and Farajtabar}]{shojaee2025illusionthinkingunderstandingstrengths}
Parshin Shojaee, Iman Mirzadeh, Keivan Alizadeh, Maxwell Horton, Samy Bengio, and Mehrdad Farajtabar. 2025.
\newblock \href {https://arxiv.org/abs/2506.06941} {The illusion of thinking: Understanding the strengths and limitations of reasoning models via the lens of problem complexity}.
\newblock \emph{Preprint}, arXiv:2506.06941.

\bibitem[{Tong et~al.(2024)Tong, Li, Wang, Wang, Teng, and Shang}]{tong2024llmslearnpreviousmistakes}
Yongqi Tong, Dawei Li, Sizhe Wang, Yujia Wang, Fei Teng, and Jingbo Shang. 2024.
\newblock \href {https://arxiv.org/abs/2403.20046} {Can llms learn from previous mistakes? investigating llms' errors to boost for reasoning}.
\newblock \emph{Preprint}, arXiv:2403.20046.

\bibitem[{Trivedi et~al.(2022)Trivedi, Balasubramanian, Khot, and Sabharwal}]{musique}
Harsh Trivedi, Niranjan Balasubramanian, Tushar Khot, and Ashish Sabharwal. 2022.
\newblock \href {https://doi.org/10.1162/tacl_a_00475} {{M}u{S}i{Q}ue: Multihop questions via single-hop question composition}.
\newblock \emph{Transactions of the Association for Computational Linguistics}, 10:533--550.

\bibitem[{Wang et~al.(2022)Wang, Wei, Schuurmans, Le, Chi, Narang, Chowdhery, and Zhou}]{wang2022self}
Xuezhi Wang, Jason Wei, Dale Schuurmans, Quoc Le, Ed~Chi, Sharan Narang, Aakanksha Chowdhery, and Denny Zhou. 2022.
\newblock Self-consistency improves chain of thought reasoning in language models.
\newblock \emph{arXiv preprint arXiv:2203.11171}.

\bibitem[{Wang et~al.(2023)Wang, Wei, Schuurmans, Le, Chi, Narang, Chowdhery, and Zhou}]{wang2023selfconsistencyimproveschainthought}
Xuezhi Wang, Jason Wei, Dale Schuurmans, Quoc Le, Ed~Chi, Sharan Narang, Aakanksha Chowdhery, and Denny Zhou. 2023.
\newblock \href {https://arxiv.org/abs/2203.11171} {Self-consistency improves chain of thought reasoning in language models}.
\newblock \emph{Preprint}, arXiv:2203.11171.

\bibitem[{Wei et~al.(2023)Wei, Wang, Schuurmans, Bosma, Ichter, Xia, Chi, Le, and Zhou}]{wei2023chainofthoughtpromptingelicitsreasoning}
Jason Wei, Xuezhi Wang, Dale Schuurmans, Maarten Bosma, Brian Ichter, Fei Xia, Ed~Chi, Quoc Le, and Denny Zhou. 2023.
\newblock \href {https://arxiv.org/abs/2201.11903} {Chain-of-thought prompting elicits reasoning in large language models}.
\newblock \emph{Preprint}, arXiv:2201.11903.

\bibitem[{Wu et~al.(2025)Wu, Yang, Wang, Okumura, and Zhang}]{wu2025cofca}
Jian Wu, Linyi Yang, Zhen Wang, Manabu Okumura, and Yue Zhang. 2025.
\newblock \href {https://openreview.net/forum?id=q2DmkZ1wVe} {Cof{CA}: A {STEP}-{WISE} counterfactual multi-hop {QA} benchmark}.
\newblock In \emph{The Thirteenth International Conference on Learning Representations}.

\bibitem[{Yan et~al.(2024)Yan, Wang, Huo, Li, Li, Su, Gao, Zhang, Xu, Chu, Zhong, Wang, Xiong, Yu, Hu, and Wen}]{yan2024errorradarbenchmarkingcomplexmathematical}
Yibo Yan, Shen Wang, Jiahao Huo, Hang Li, Boyan Li, Jiamin Su, Xiong Gao, Yi-Fan Zhang, Tianlong Xu, Zhendong Chu, Aoxiao Zhong, Kun Wang, Hui Xiong, Philip~S. Yu, Xuming Hu, and Qingsong Wen. 2024.
\newblock \href {https://arxiv.org/abs/2410.04509} {Errorradar: Benchmarking complex mathematical reasoning of multimodal large language models via error detection}.
\newblock \emph{Preprint}, arXiv:2410.04509.

\bibitem[{Yang et~al.(2018)Yang, Qi, Zhang, Bengio, Cohen, Salakhutdinov, and Manning}]{hotpotqa}
Zhilin Yang, Peng Qi, Saizheng Zhang, Yoshua Bengio, William~W. Cohen, Ruslan Salakhutdinov, and Christopher~D. Manning. 2018.
\newblock \href {http://arxiv.org/abs/1809.09600} {Hotpotqa: {A} dataset for diverse, explainable multi-hop question answering}.
\newblock \emph{CoRR}.

\bibitem[{Zheng et~al.(2024)Zheng, Zhang, Zhang, Lin, Lu, Yu, Liu, Zhou, and Lin}]{zheng2024processbenchidentifyingprocesserrors}
Chujie Zheng, Zhenru Zhang, Beichen Zhang, Runji Lin, Keming Lu, Bowen Yu, Dayiheng Liu, Jingren Zhou, and Junyang Lin. 2024.
\newblock \href {https://arxiv.org/abs/2412.06559} {Processbench: Identifying process errors in mathematical reasoning}.
\newblock \emph{Preprint}, arXiv:2412.06559.

\end{thebibliography}

\appendix
\section{Appendix A}
\begin{table*}[ht]
\small
\centering
\begin{tabularx}{\textwidth}{p{2.8cm} p{3.5cm} Y}
\toprule
\textbf{Question Type} & \textbf{Bridge Entity / Reasoning} & \textbf{Example} \\
\midrule

\textbf{Compositional}: \\Requires chaining intermediate
entities. & Bridge: \textcolor{blue}{Versus} & 
\textbf{Doc 1:} \textcolor{blue}{Versus} (Versace) is the diffusion line of Italian..., a gift by the founder \textcolor{blue}{Gianni Versace}.\par
\textbf{Doc 2:} \textcolor{blue}{Gianni Versace} was shot and killed outside...\par
\textbf{Question:} Why did the founder of \textcolor{blue}{Versus} die? \\

\midrule

\textbf{Inference}: \\Demands implicit reasoning via unstated bridge facts & Bridge: \textcolor{blue}{Grandchild} & 
\textbf{Doc 1:} \textcolor{blue}{Dambar Shah} was the father of \textcolor{blue}{Krishna Shah}...\par
\textbf{Doc 2:} \textcolor{blue}{Krishna Shah} was the father of \textcolor{blue}{Rudra Shah}...\par
\textbf{Question:} Who is the \textcolor{blue}{grandchild} of \textcolor{blue}{Dambar Shah}? \\

\midrule

\textbf{Comparison}: \\Involves comparing attributes across entities & Compare: \textcolor{blue}{Age} of Persons & 
\textbf{Doc 1:} \textcolor{blue}{Theodor Haecker} (1879–1945) was a...\par
\textbf{Doc 2:} \textcolor{blue}{Harry Vaughan Watkins} (1875–1945) was a...\par
\textbf{Question:} Who lived longer, \textcolor{blue}{Theodor Haecker} or \textcolor{blue}{Harry Vaughan Watkins}? \\

\midrule

\textbf{Bridge Comparison}: \\Combines inference followed by comparison & Bridge: Directors' \textcolor{blue}{Nationality} & 
\textbf{Doc 1:} \textcolor{blue}{FAQ: Frequently Asked Questions} directed by \textcolor{blue}{Carlos Atanes}...\par
\textbf{Doc 2:} \textcolor{blue}{The Big Money} directed by \textcolor{blue}{John Paddy Carstairs}...\par
\textbf{Doc 3:} \textcolor{blue}{Carlos Atanes} is a Spanish film director.\par
\textbf{Doc 4:} \textcolor{blue}{John Paddy Carstairs} was a British film director.\par
\textbf{Question:} Are both directors of \textcolor{blue}{FAQ: Frequently Asked Questions} and \textcolor{blue}{The Big Money} from the same country? \\

\bottomrule
\end{tabularx}
\caption{Examples of Different Question Types with Highlighted Entities.}
\label{Appendix:question-types}
\end{table*}

\section*{Stage 1 Categories (Coarse Taxonomy)}
\begin{enumerate}
    \item \textbf{Effective reasoning:} The model performs all required reasoning steps and correctly answers the question. The explanation is concise, coherent, and logically complete.
    
    \item \textbf{Underthinking:} The model provides insufficient reasoning, skipping essential steps or offering vague justifications. The response may appear shallow or overly brief, regardless of answer correctness.
    
    \item \textbf{Overthinking:} The model introduces excessive or tangential reasoning, often by exploring irrelevant paths or repeating information. This may include unnecessary document traversal or redundant entity comparisons.
    
    \item \textbf{Faulty reasoning:} The reasoning chain is logically flawed or factually incorrect. This may involve wrong inference, unsupported claims, or internal contradictions, even if the structure appears complete.
\end{enumerate}

\section*{Stage 2 Categories (Structured Taxonomy)}

Let $N_{\text{model}}$ denote the number of reasoning steps predicted by the model, and $N_{\text{gold}}$ denote the number of hops required according to the gold standard.

\begin{enumerate}
    \item \textbf{Category 1: $N_{\text{model}} = N_{\text{gold}}$; all hops correct; final answer correct} \\ The model follows the required inference path, makes all correct hops, and provides the correct final answer.

    \item \textbf{Category 2: $N_{\text{model}} = N_{\text{gold}}$; all hops correct; final answer incorrect} \\ The reasoning path is structurally correct, but the final answer is wrong due to errors in aggregation or conclusion.

    \item \textbf{Category 3: $N_{\text{model}} = N_{\text{gold}}$; one or more hops incorrect/hallucinated} \\ The hop count matches, but one or more steps are logically or factually incorrect. The final answer may or may not be correct.

    \item \textbf{Category 4: $N_{\text{model}} < N_{\text{gold}}$; all predicted hops correct; final answer incorrect} \\ The model correctly predicts a subset of the required hops but misses key steps, leading to an incorrect answer.

    \item \textbf{Category 5: $N_{\text{model}} < N_{\text{gold}}$; all predicted hops correct; final answer correct (shortcut)} \\ The model answers correctly using a valid but incomplete subset of required reasoning hops. A shortcut was taken.

    \item \textbf{Category 6: $N_{\text{model}} < N_{\text{gold}}$; one or more hops incorrect/hallucinated} \\ The model generates fewer hops than required, with some being inaccurate or irrelevant. The chain is both incomplete and partially flawed.

    \item \textbf{Category 7: $N_{\text{model}} > N_{\text{gold}}$; irrelevant hops after gold path (trailing overthinking)} \\ After attempting the required reasoning, the model continues with superfluous or irrelevant steps, leading to overgeneration.

    \item \textbf{Category 8: $N_{\text{model}} > N_{\text{gold}}$; irrelevant or hallucinated hops before/interleaved} \\ Irrelevant hops occur early in the reasoning process or are interleaved with required steps, disrupting logical progression.

    \item \textbf{Category 9: $N_{\text{model}} = 0$} \\ No reasoning path is shown; the model outputs an answer directly without generating any hops.

    \item \textbf{Category 10: Question misinterpretation} \\ The reasoning chain reflects a misunderstanding of the question, regardless of hop count or structural form.
\end{enumerate}

\section*{Dataset Details}

\subsection{Setup}

\begin{table*}[ht]
\centering
\small
\begin{tabular}{p{3cm} p{4cm} p{1cm} p{2cm}}
\toprule
\textbf{Dataset} & \textbf{Question Types} & \textbf{\#Hops} & \textbf{Difficulty} \\
\midrule
HotpotQA & Composition, Comparison & 2 & Easy–Medium \\
2WikiMultiHopQA & Composition, Comparison, Inference, Bridge Comparison & 2, 4 & Medium–Hard \\
MuSiQue & Composition, Inference & 2, 3, 4 & Hard \\
\bottomrule
\end{tabular}
\caption{Comparison of datasets in terms of question types, hop complexity, and difficulty level.}
\label{tab:dataset-stats}
\end{table*}

\begin{itemize}
\item \textbf{2WikiMultiHopQA}~\cite{ho2020constructingmultihopqadataset}:  
Includes 10 Wikipedia paragraphs per question, retrieved from structured and unstructured sources. Each instance provides gold reasoning paths and supporting facts to ensure multi-hop inference.
\item \textbf{HotpotQA}~\cite{hotpotqa}:  
Each question is paired with two gold documents and eight distractors (10 in total). Designed to test both answer accuracy and reasoning transparency, including bridge and comparison questions with annotated supporting sentences.
\item \textbf{MuSiQue}~\cite{musique}: Presents 20-document contexts per question. Constructed to reduce shortcut-based reasoning by enforcing sub-question dependencies and including challenging unanswerable distractors.
\end{itemize}

\begin{figure*}[ht]
    \centering
    \begin{minipage}{\textwidth}
        \centering
        \includegraphics[width=0.8\textwidth]{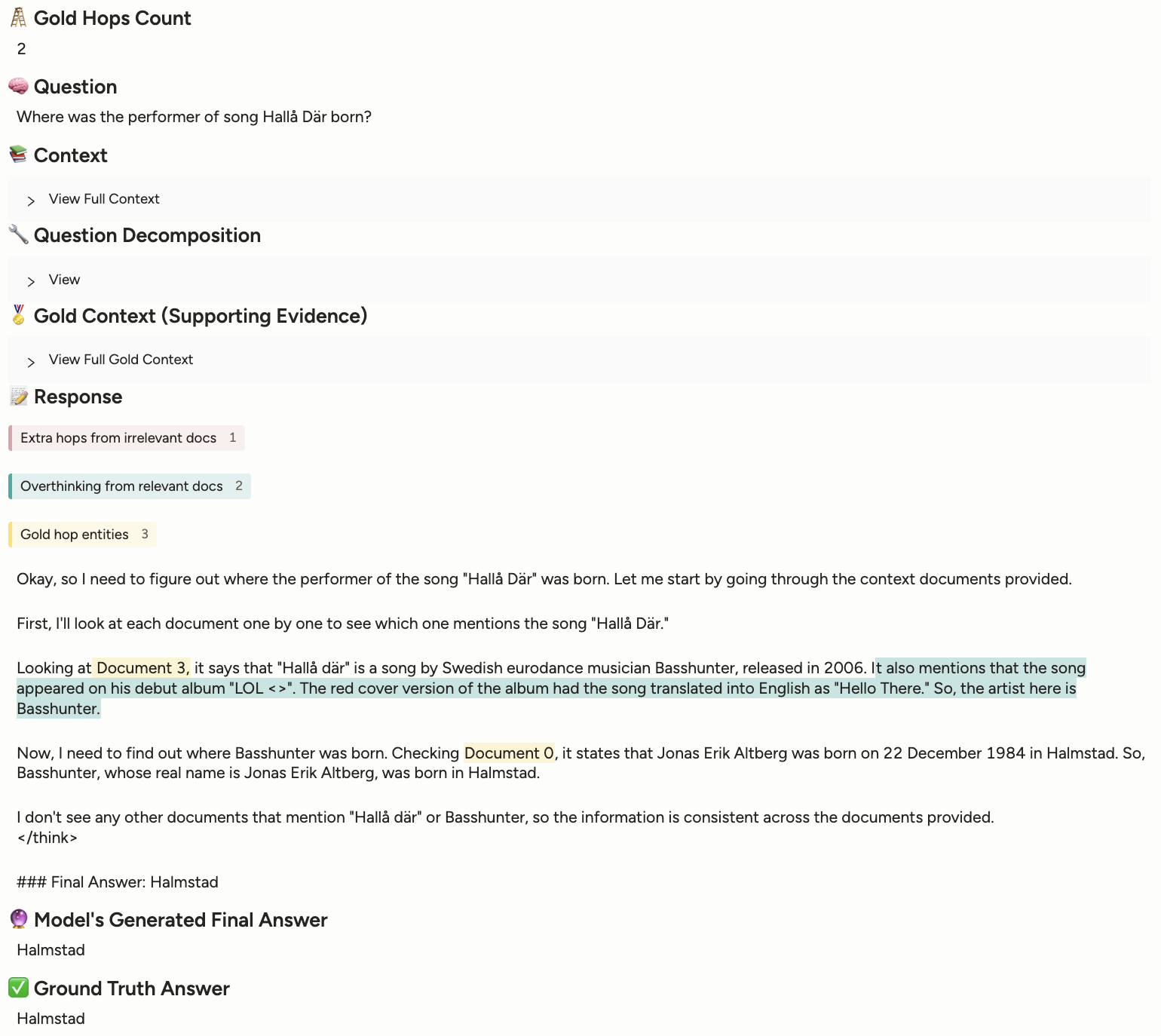}
        \caption{Example of Input given to annotators}
        \label{fig:interface1}
    \end{minipage}
\end{figure*}

\begin{figure*}[ht]
    \centering
    \begin{minipage}{\textwidth}
        \centering
        \includegraphics[width=0.8\textwidth]{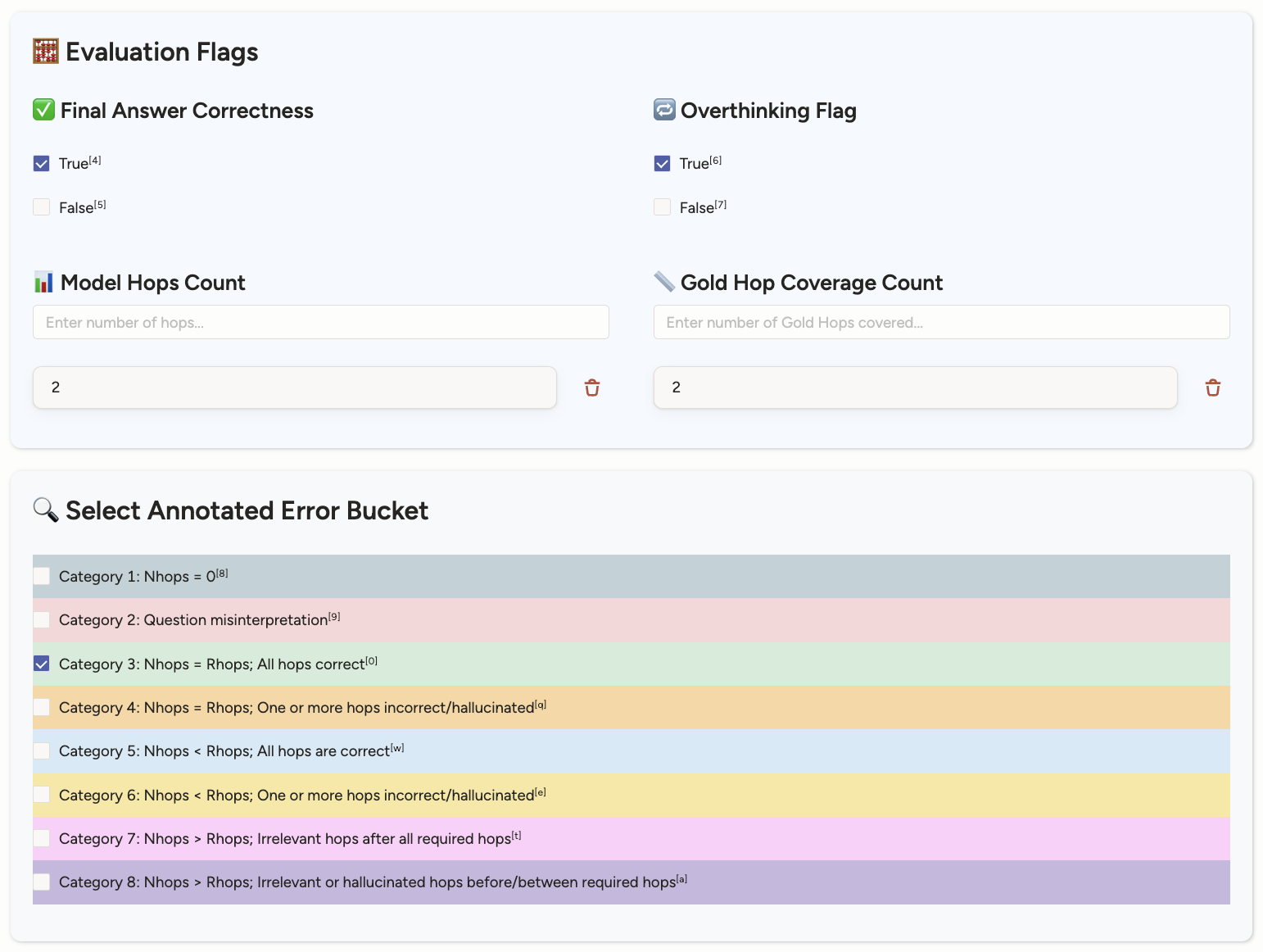}
        \caption{Example of Output labeled by the annotators}
        \label{fig:interface2}
    \end{minipage}
\end{figure*}


\section{Additional Figures}
\label{appendix:full-results}


\begin{figure*}[htbp]
    \centering

    \begin{subfigure}[b]{0.9\textwidth}
        \centering
        \includegraphics[width=\linewidth]{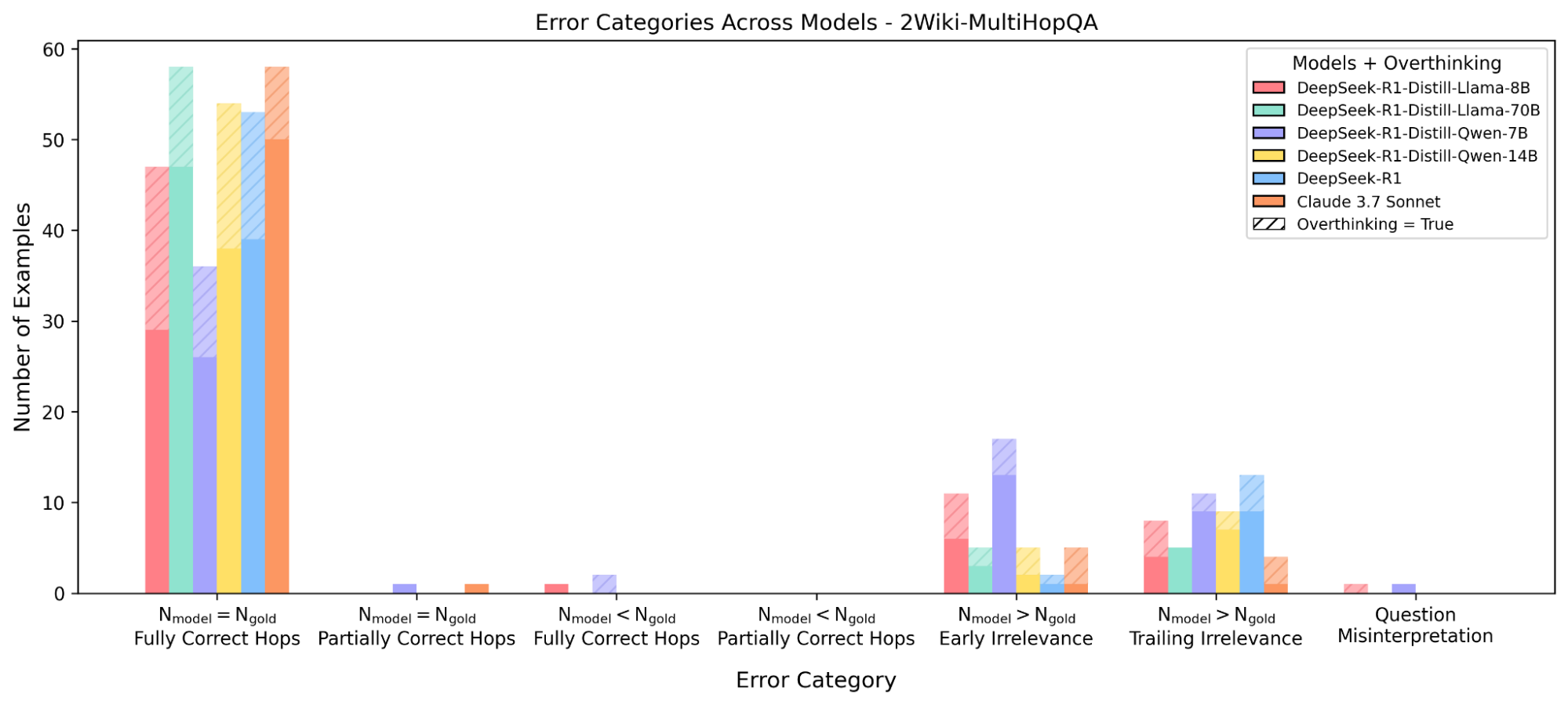}
        \caption{2Wiki}
        \label{fig:reasoning-errors-2wiki}
    \end{subfigure}

    \vspace{1em}

    \begin{subfigure}[b]{0.9\textwidth}
        \centering
        \includegraphics[width=\linewidth]{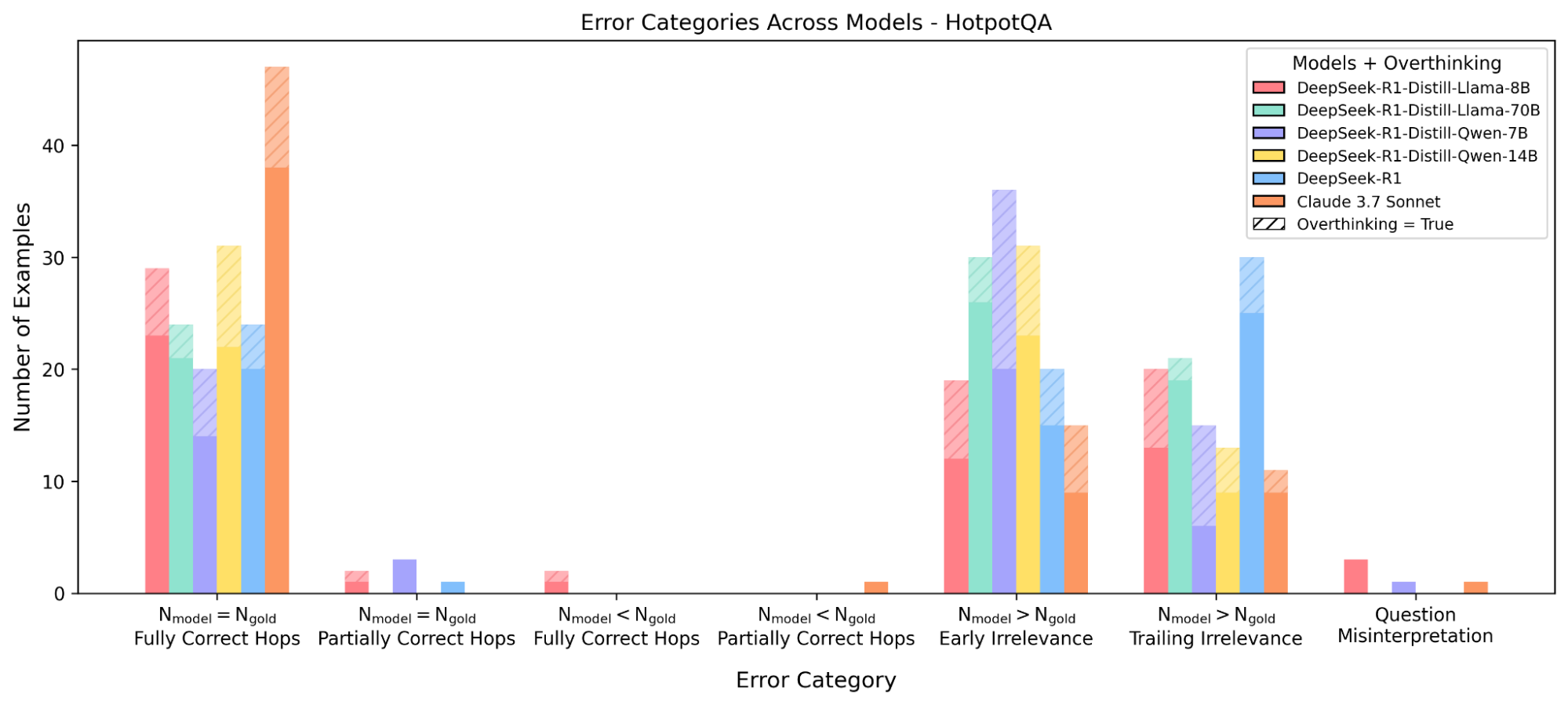}
        \caption{HotpotQA}
        \label{fig:reasoning-errors-hotpot}
    \end{subfigure}

    \vspace{1em}

    \begin{subfigure}[b]{0.9\textwidth}
        \centering
        \includegraphics[width=\linewidth]{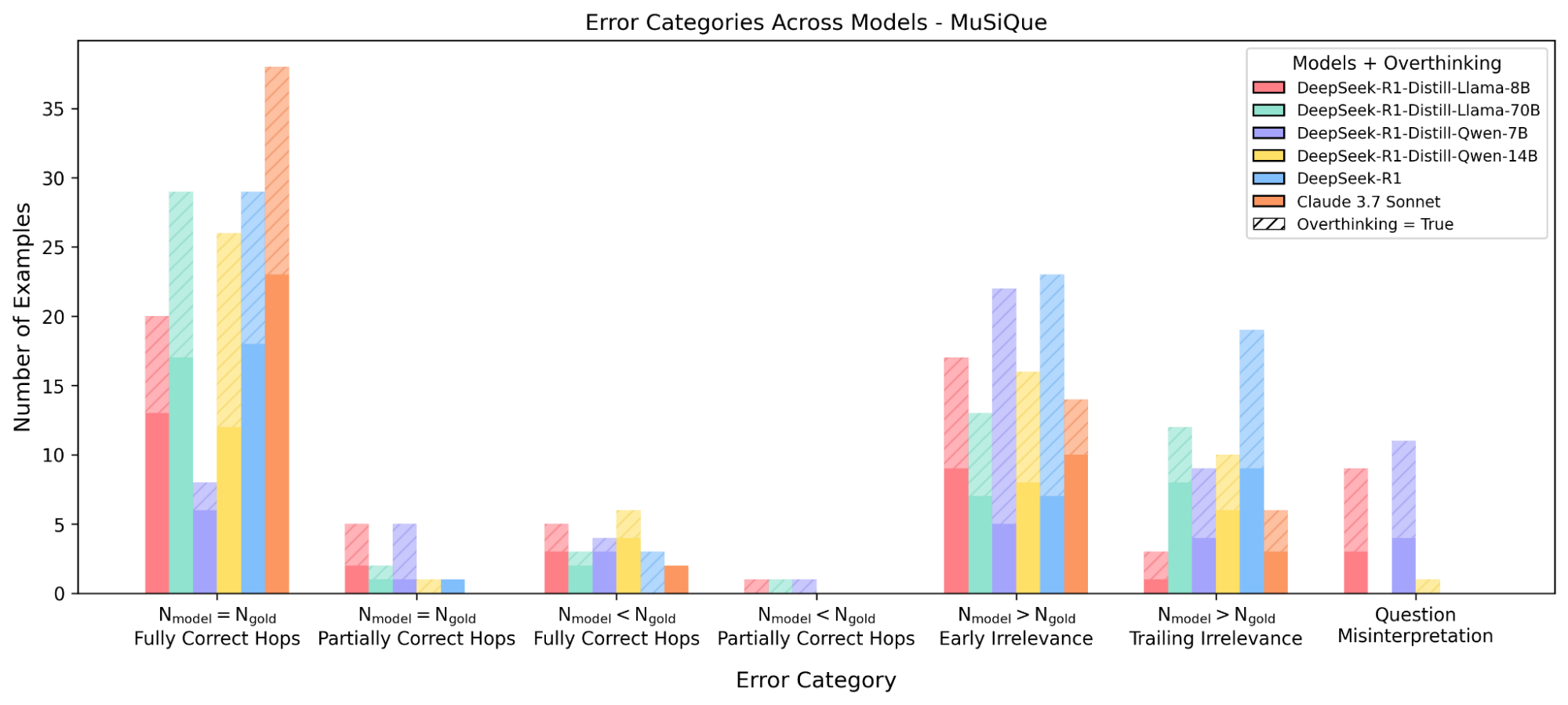}
        \caption{MuSiQue}
        \label{fig:reasoning-errors-mus}
    \end{subfigure}

    \caption{Distribution of reasoning error types across datasets. (a) \textsc{2Wiki}, (b) \textsc{HotpotQA}, (c) \textsc{MuSiQue}.}
    \label{fig:reasoning-errors-all}
\end{figure*}


\begin{figure*}[htbp]
    \centering

    \begin{subfigure}[b]{0.9\textwidth}
        \centering
        \includegraphics[width=\linewidth]{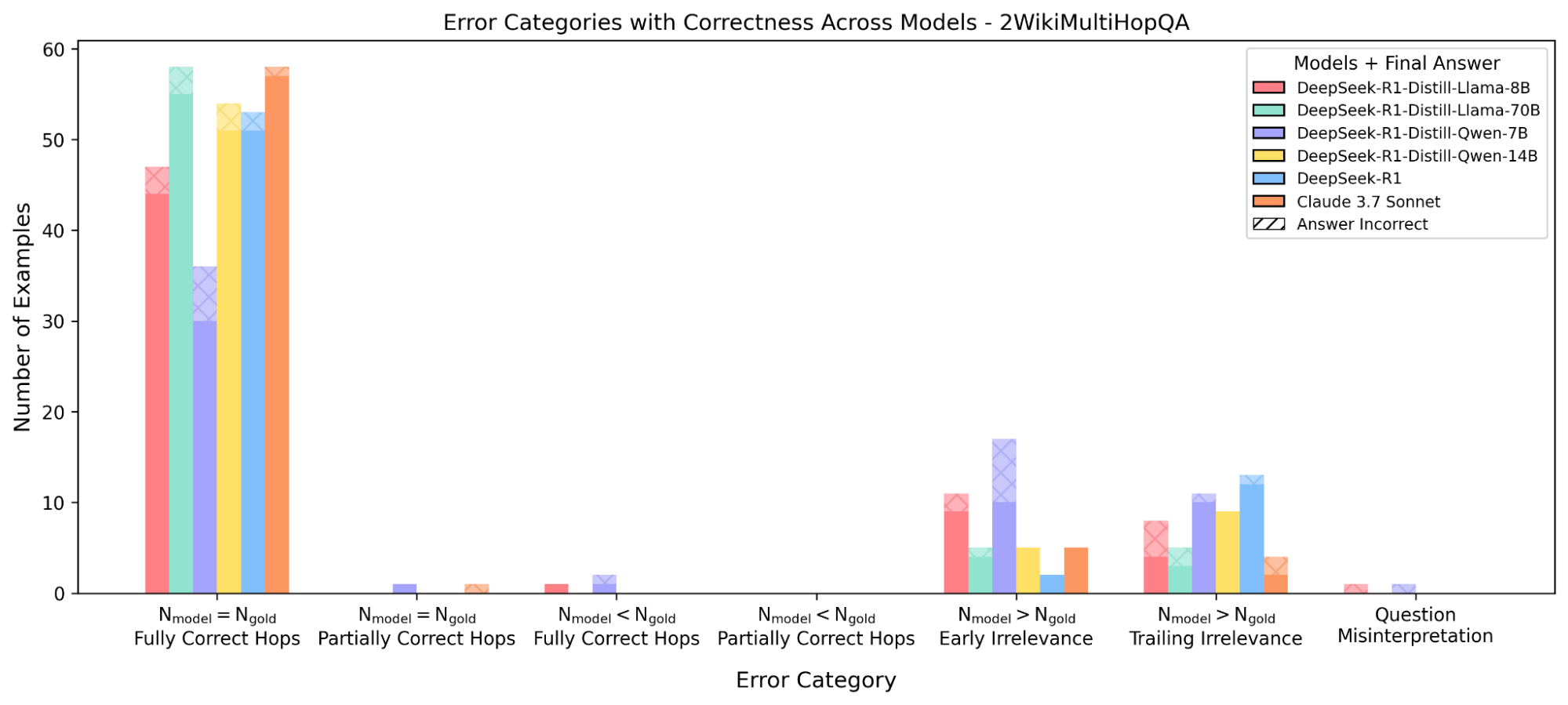}
        \caption{2Wiki}
        \label{fig:answer-correctness-2wiki}
    \end{subfigure}

    \vspace{1em}

    \begin{subfigure}[b]{0.9\textwidth}
        \centering
        \includegraphics[width=\linewidth]{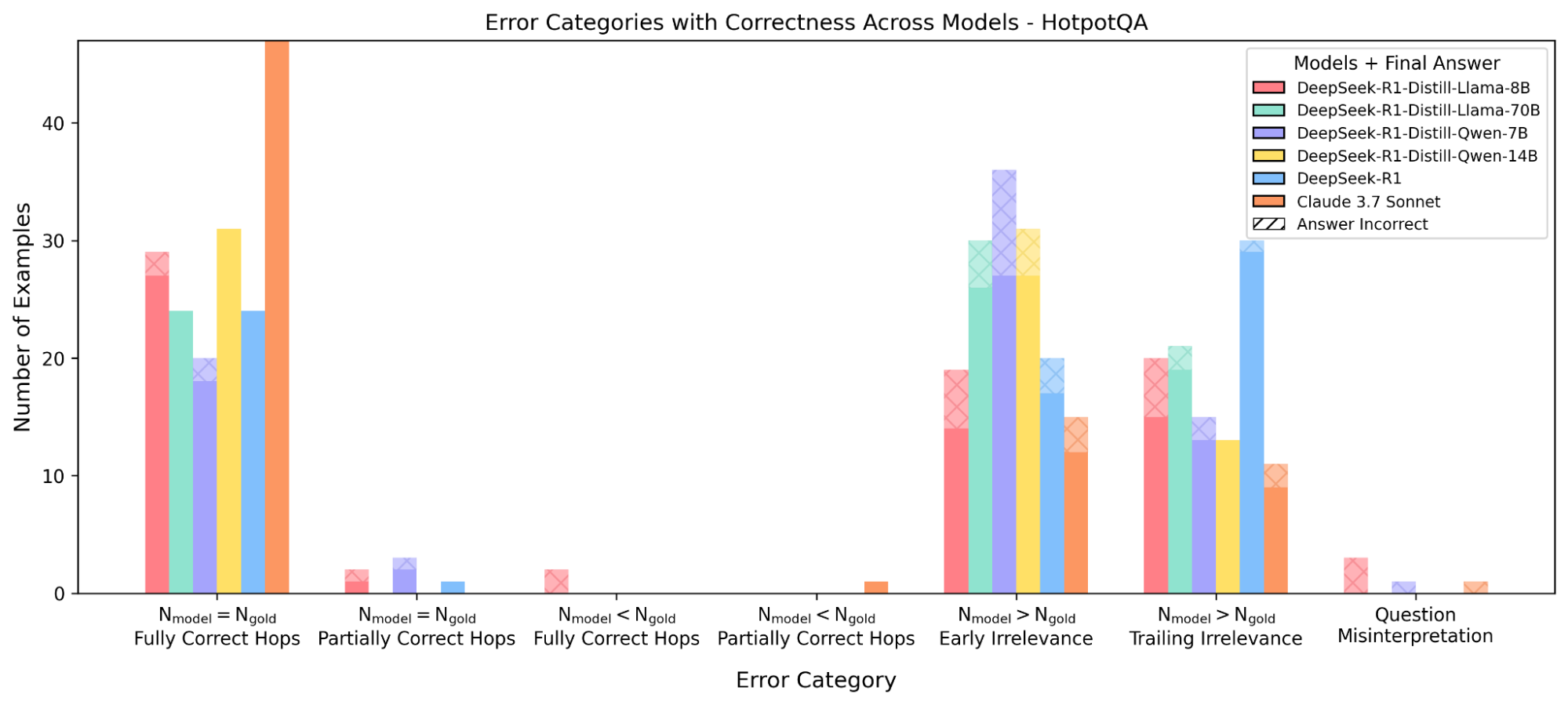}
        \caption{HotpotQA}
        \label{fig:answer-correctness-hotpot}
    \end{subfigure}

    \begin{subfigure}[b]{0.9\textwidth}
        \centering
        \includegraphics[width=\linewidth]{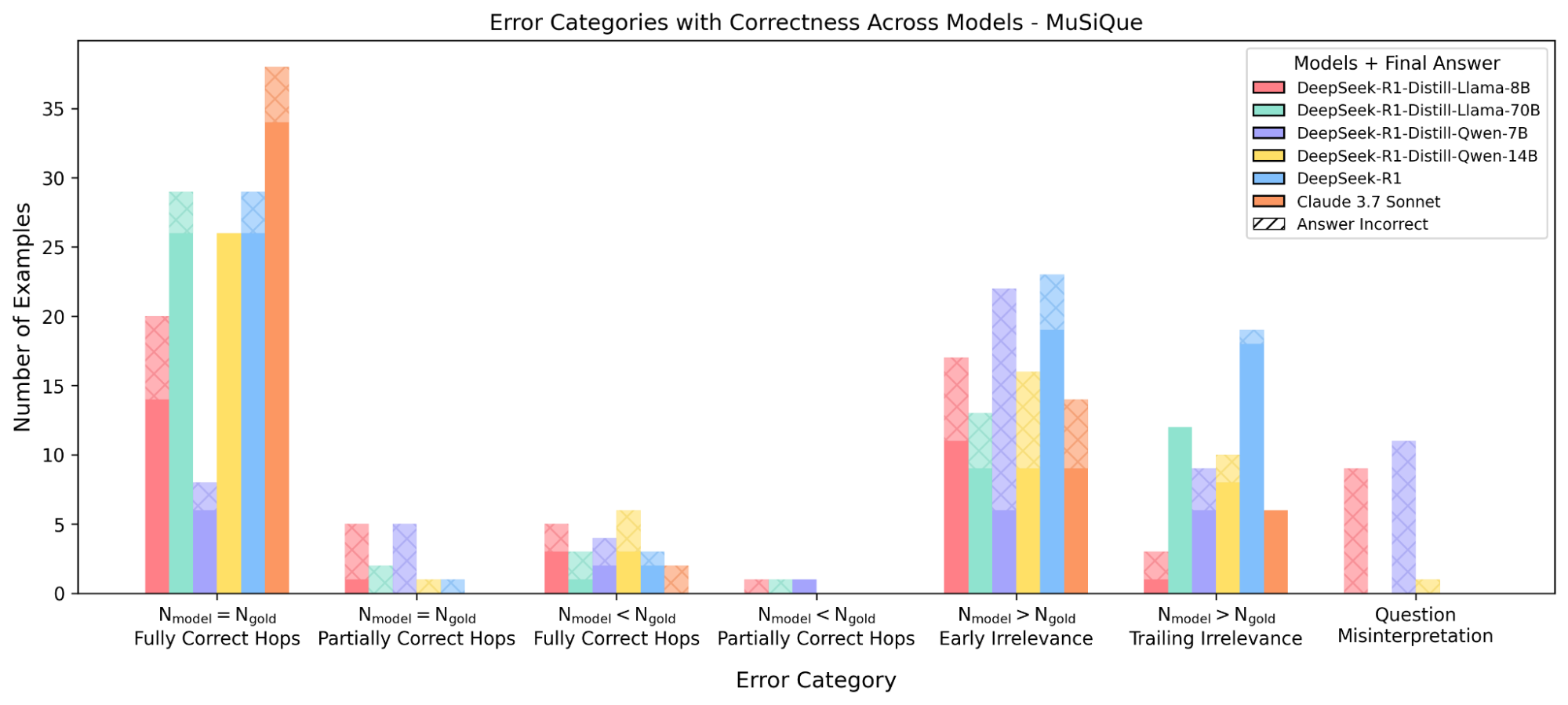}
        \caption{MuSiQue}
        \label{fig:answer-correctness-mus}
    \end{subfigure}

    \caption{Answer correctness breakdown by reasoning category across datasets. (a) \textsc{2Wiki}, (b) \textsc{HotpotQA}, (c) \textsc{MuSiQue}.}
    \label{fig:answer-correctness-all}
\end{figure*}


\begin{figure*}[htbp]
    \centering

    \begin{subfigure}[b]{0.9\textwidth}
        \centering
        \includegraphics[width=\textwidth]{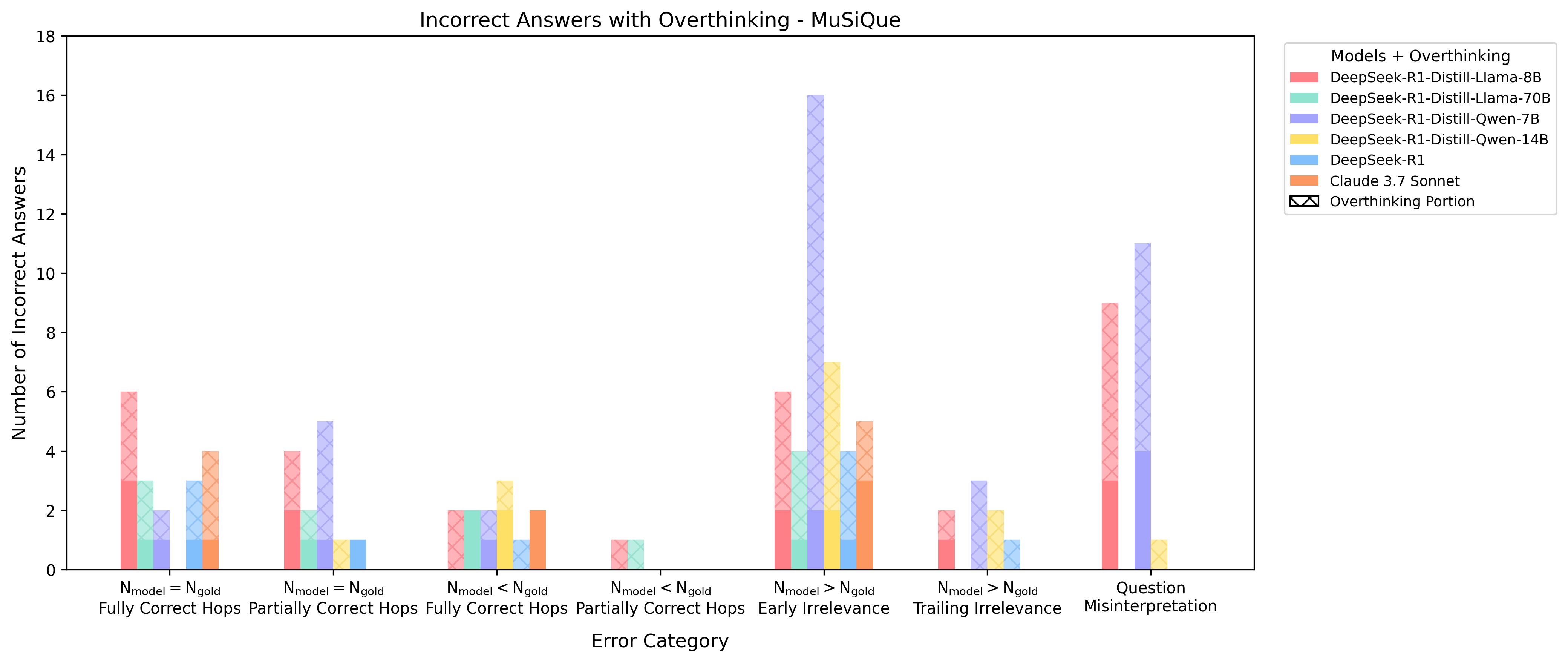}
        \caption{MuSiQue}
        \label{fig:answer-incorrectness-musique-overthink}
    \end{subfigure}
    
    \vspace{0.8em}  

    \begin{subfigure}[b]{0.9\textwidth}
        \centering
        \includegraphics[width=\textwidth]{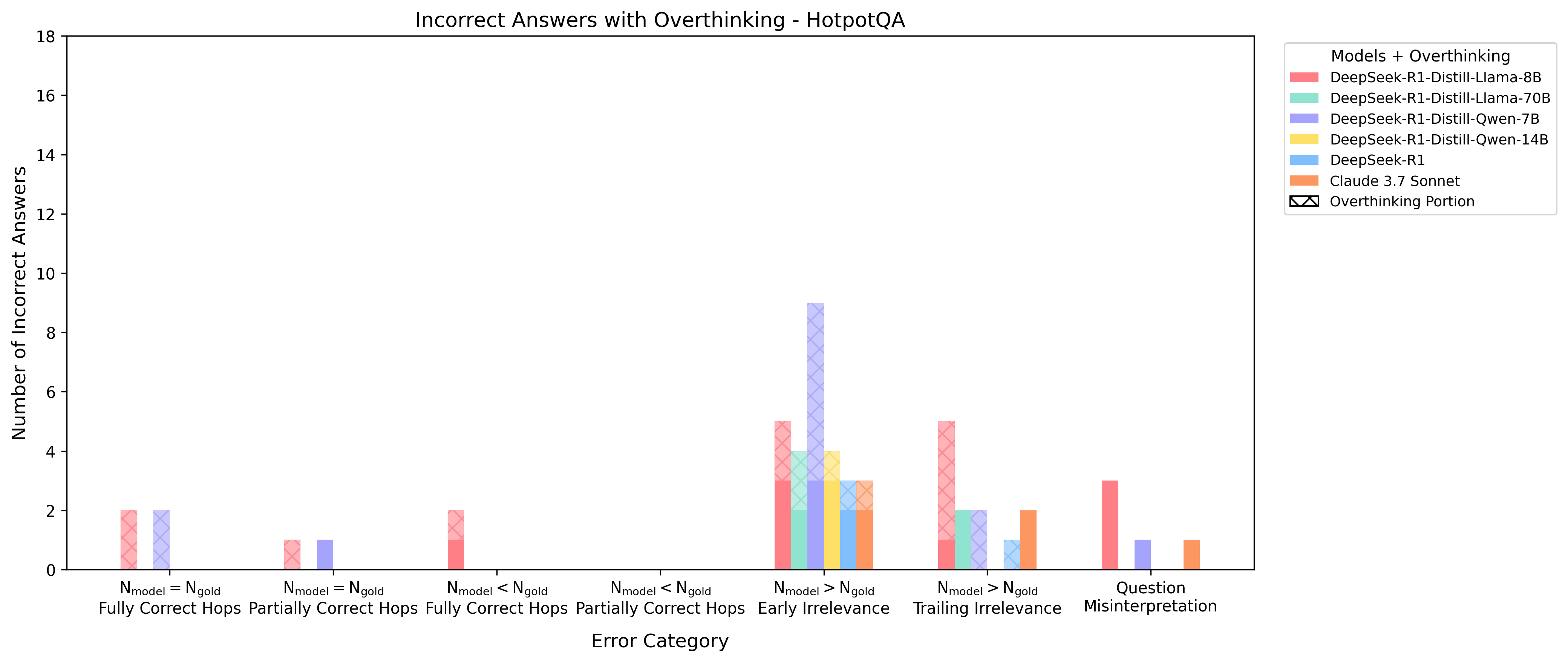}
        \caption{HotpotQA}
        \label{fig:answer-incorrectness-hotpot-overthink}
    \end{subfigure}

    \vspace{0.8em}  

    \begin{subfigure}[b]{0.9\textwidth}
        \centering
        \includegraphics[width=\textwidth]{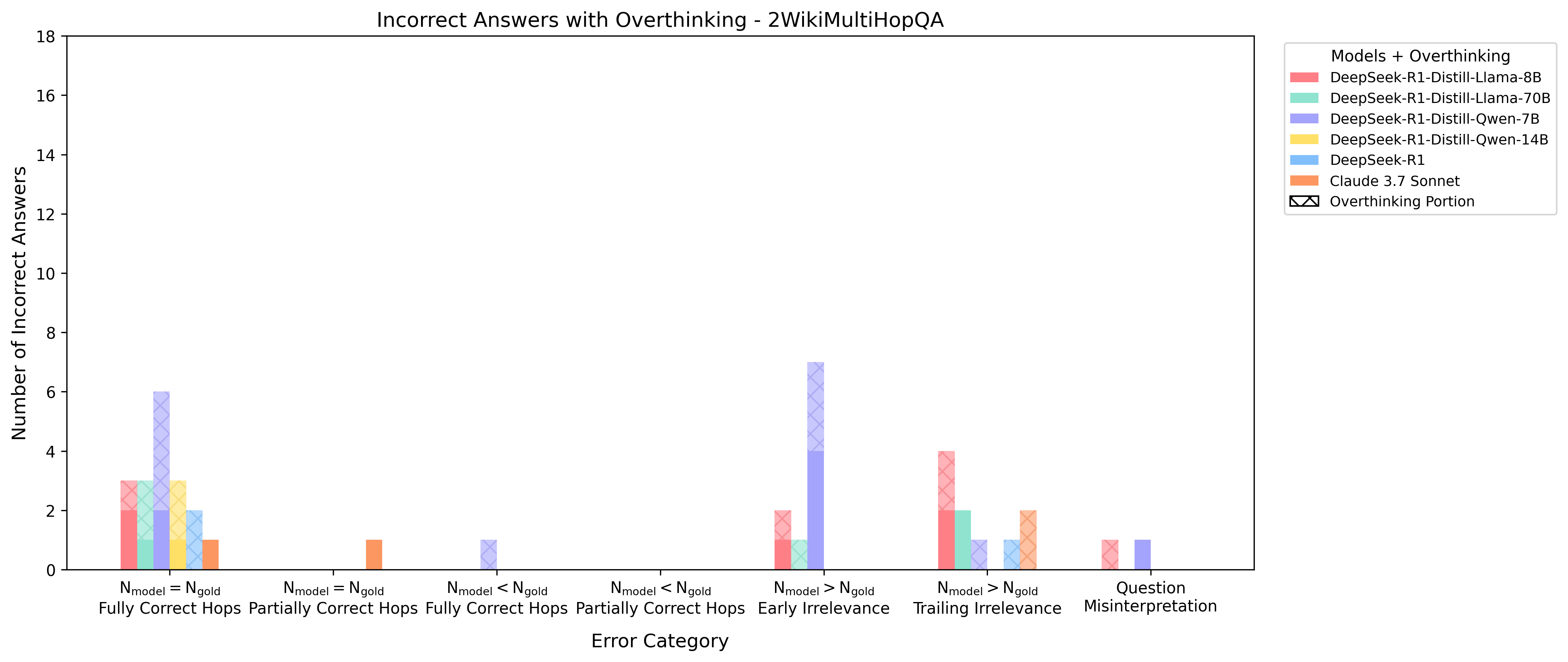}
        \caption{2Wiki}
        \label{fig:answer-incorrectness-2wiki-overthink}
    \end{subfigure}

    \caption{Overthinking Trends with Answer Incorrectness across Datasets. (a) MuSiQue, (b) HotpotQA, and (c) 2Wiki.}
    \label{fig:answer-incorrectness-overthink-all}
\end{figure*}


\begin{figure*}[htbp]
    \centering

    \begin{subfigure}[b]{0.45\textwidth}
        \centering
        \includegraphics[width=\linewidth]{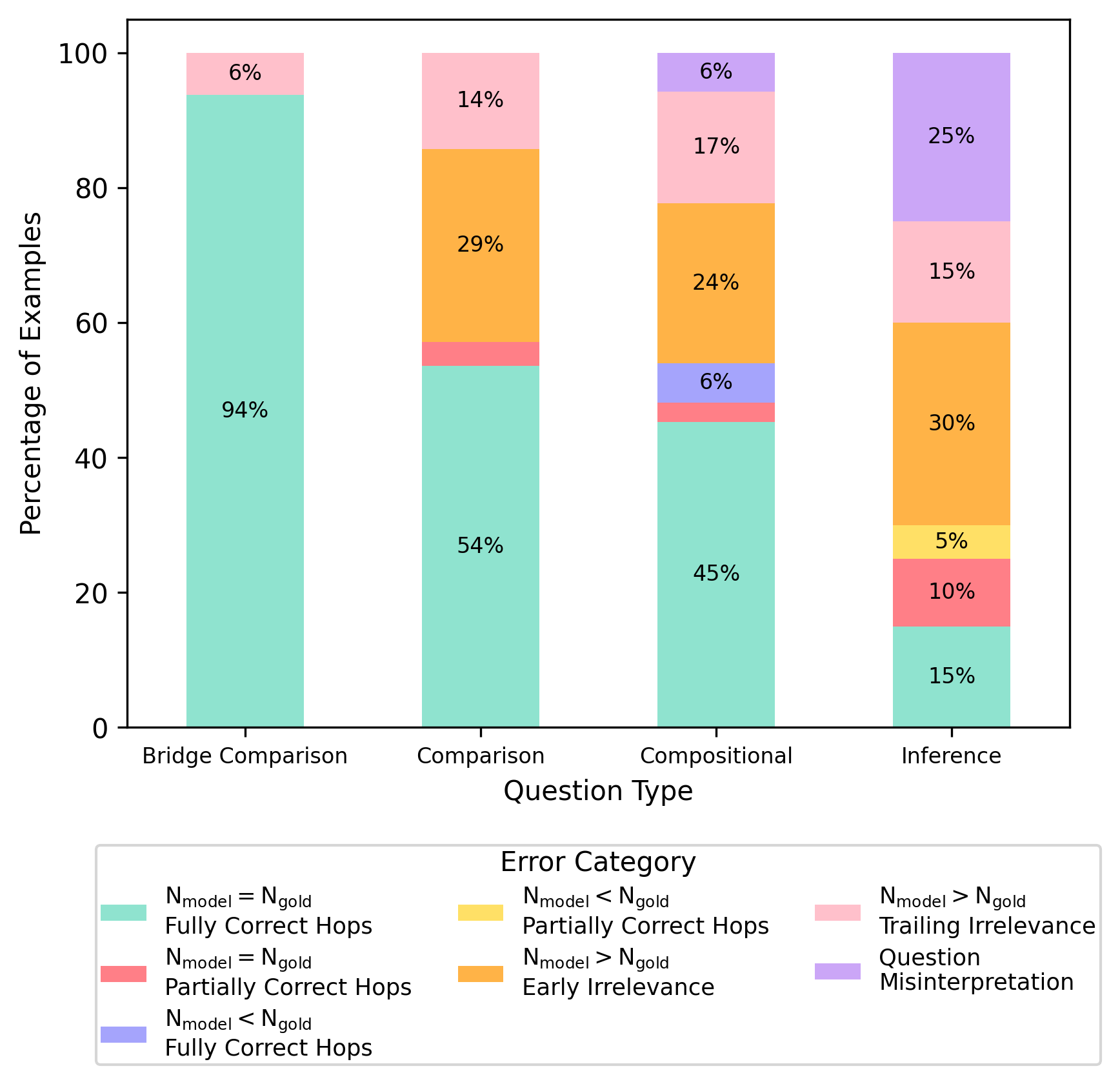}
        \caption{LLaMA-8B (Distill)}
        \label{fig:llama70b_qtypes}
    \end{subfigure}
    \hfill
    \begin{subfigure}[b]{0.45\textwidth}
        \centering
        \includegraphics[width=\linewidth]{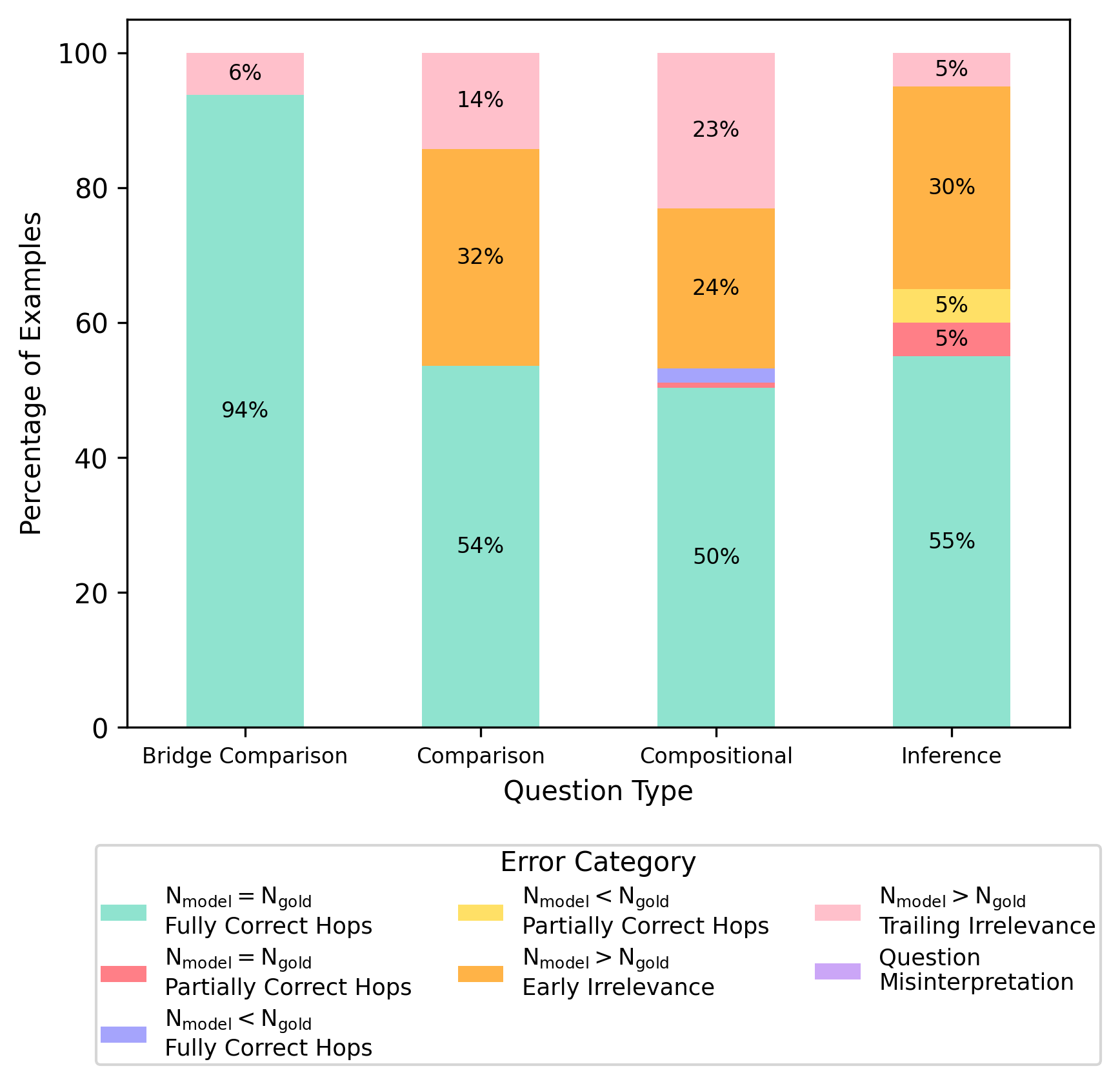}
        \caption{LLaMA-70B (Distill)}
        \label{fig:LLama70_qtypes}
    \end{subfigure}

    \begin{subfigure}[b]{0.45\textwidth}
        \centering
        \includegraphics[width=\linewidth]{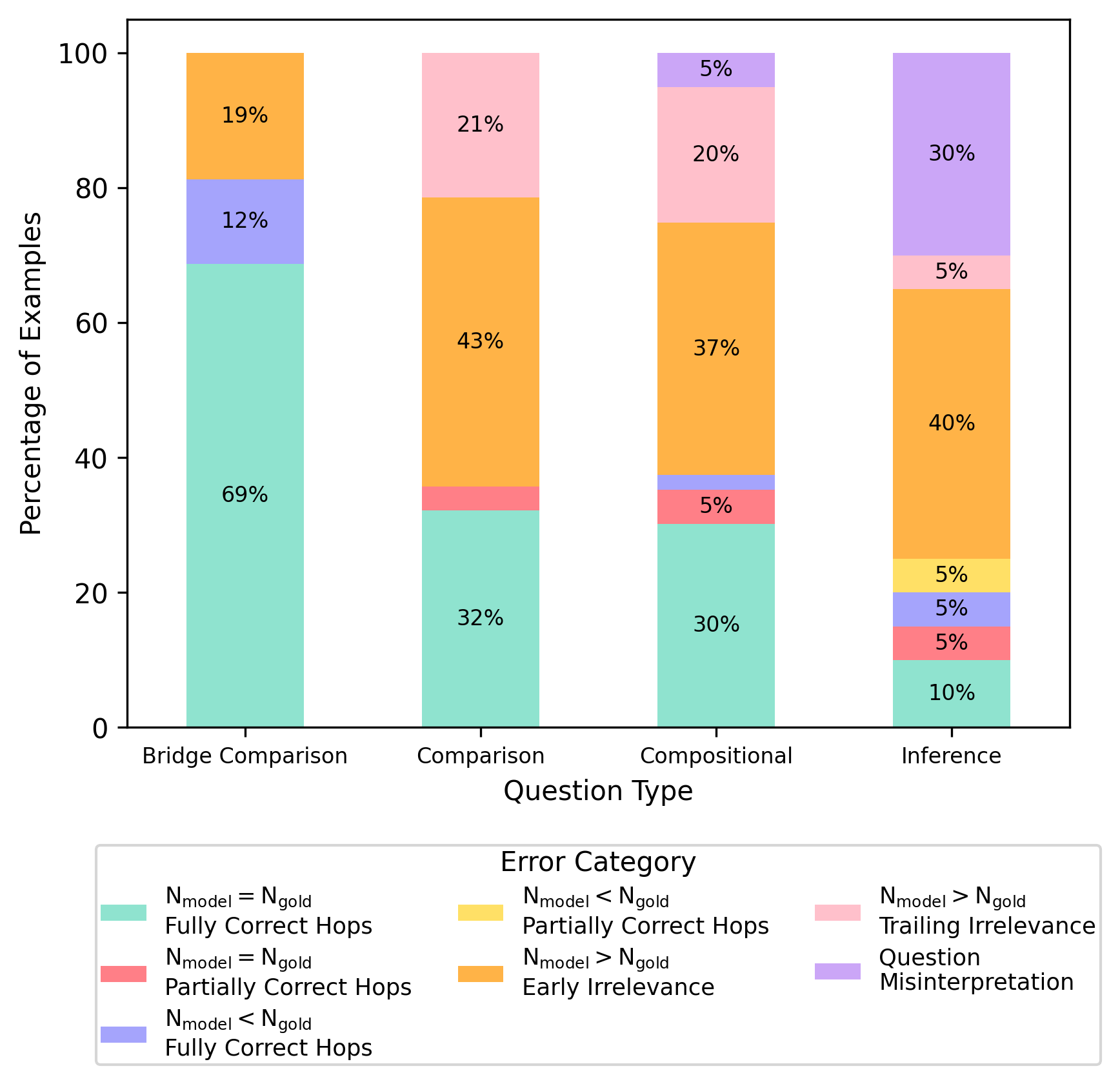}
        \caption{Qwen-7B (Distill)}
        \label{fig:qwen7b_qtypes}
    \end{subfigure}
    \hfill
    \begin{subfigure}[b]{0.45\textwidth}
        \centering
        \includegraphics[width=\linewidth]{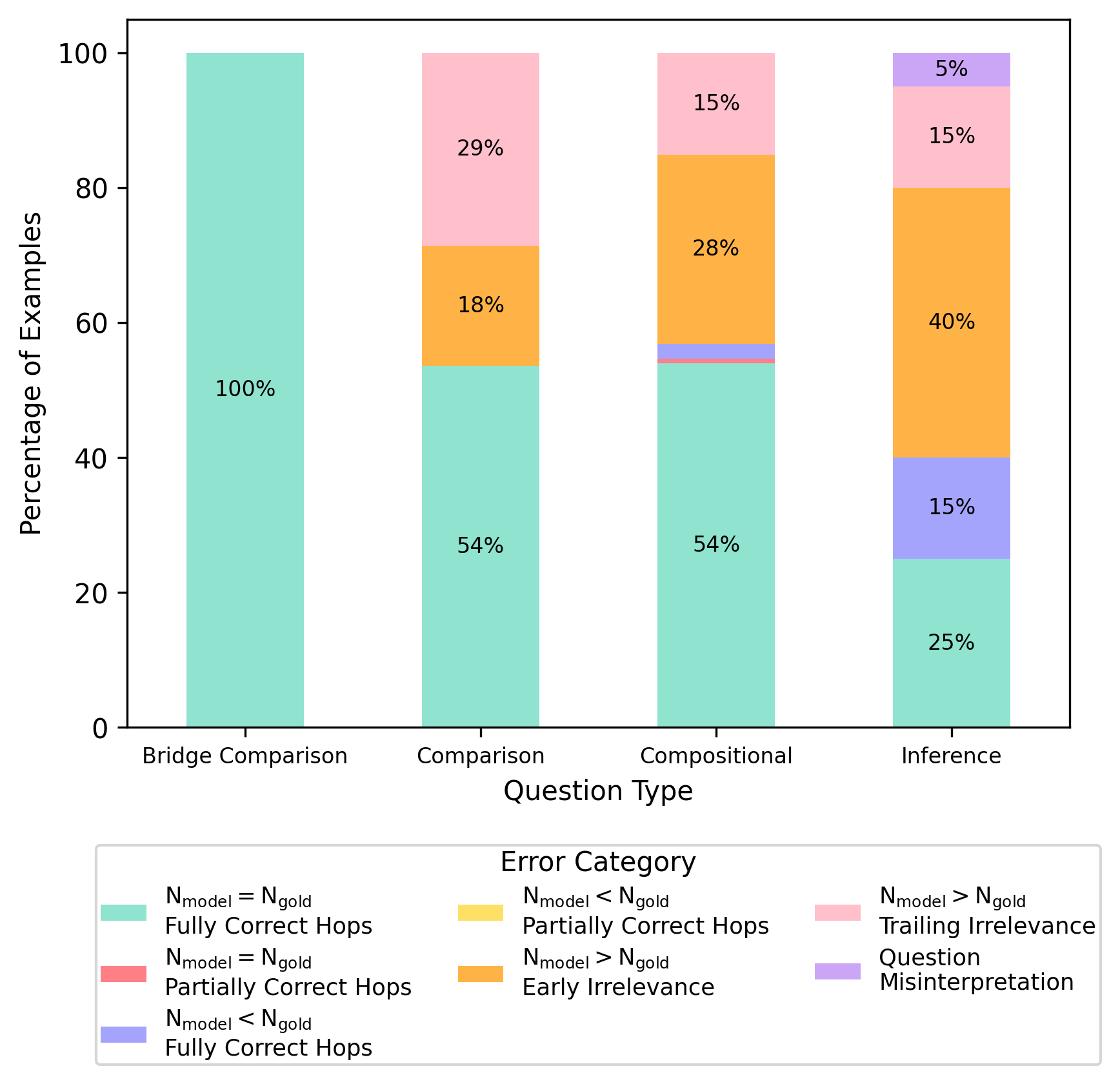}
        \caption{Qwen-14B (Distill)}
        \label{fig:qwen14b_qtypes}
    \end{subfigure}

    \begin{subfigure}[b]{0.45\textwidth}
        \centering
        \includegraphics[width=\linewidth]{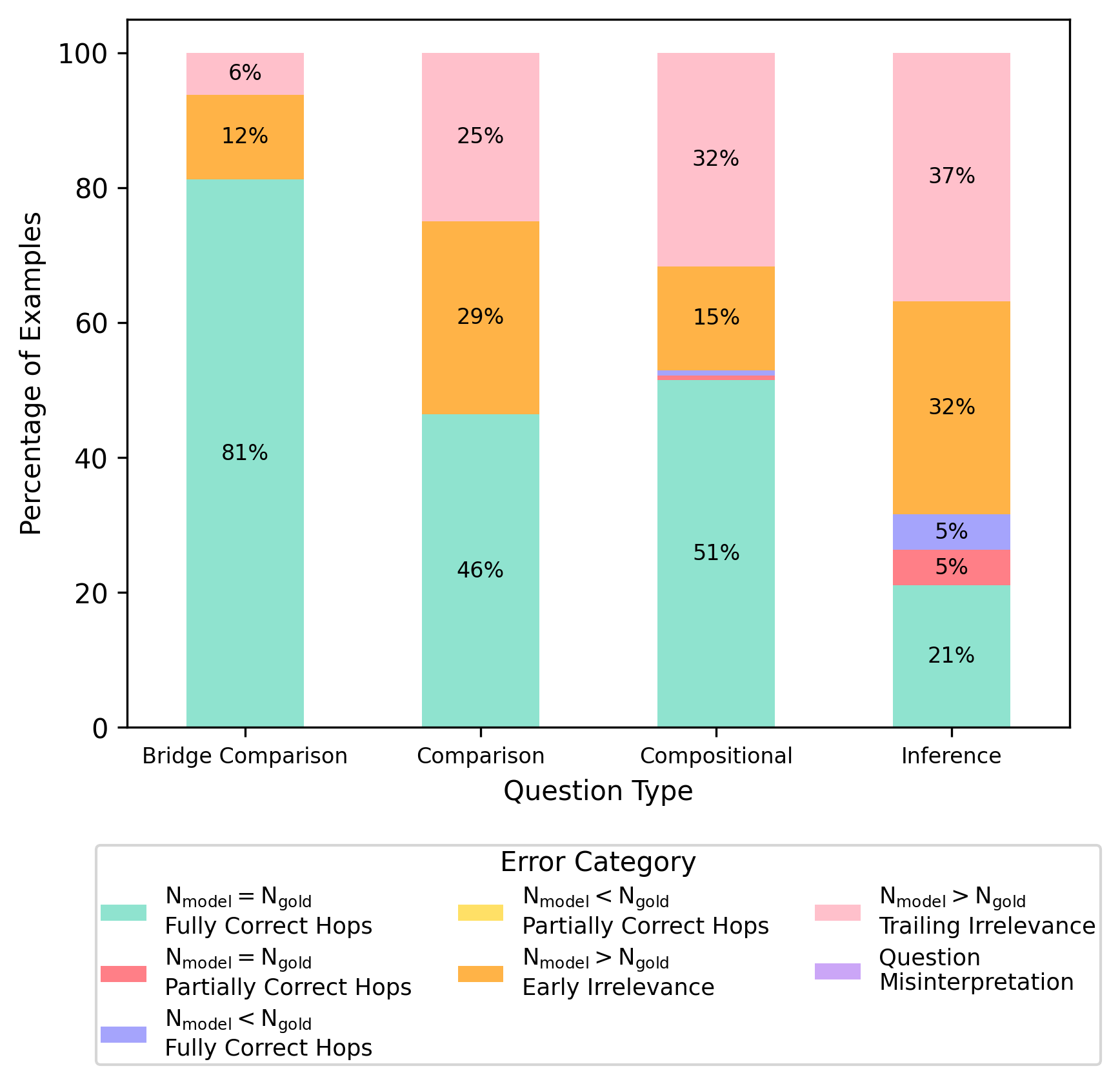}
        \caption{DeepSeek-R1}
        \label{fig:deepseekr1_qtypes}
    \end{subfigure}
    \hfill
    \begin{subfigure}[b]{0.45\textwidth}
        \centering
        \includegraphics[width=\linewidth]{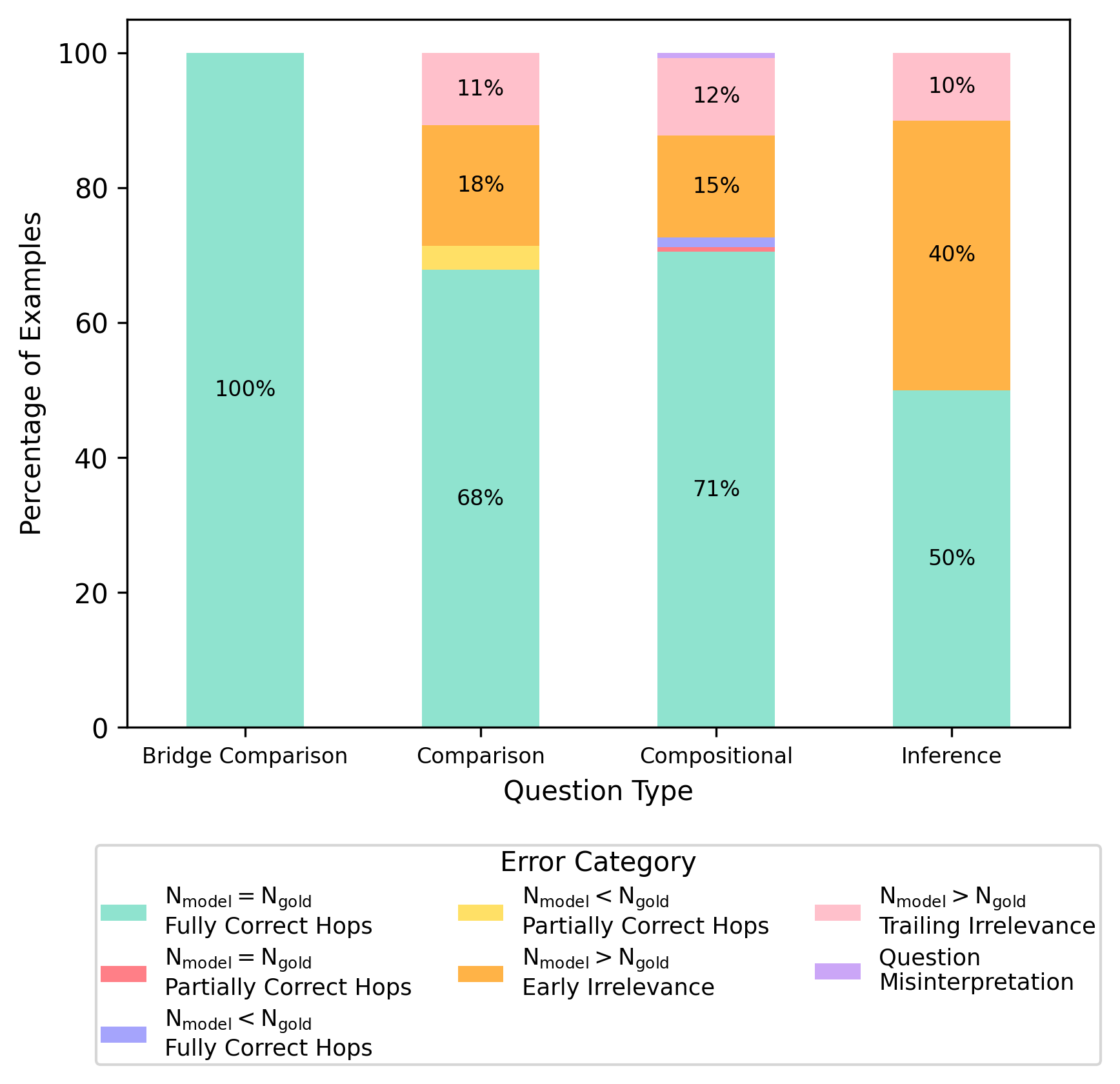}
        \caption{Claude 3 Sonnet}
        \label{fig:claude_qtypes}
    \end{subfigure}

    \caption{Distribution of reasoning error types across question types for six models. Each subfigure shows model-specific trends in how question type impacts reasoning errors.}
    \label{fig:qtypes_error_all}
\end{figure*}

\begin{figure*}[htbp]
    \centering

    \begin{subfigure}[b]{0.48\textwidth}
        \centering
        \includegraphics[width=\linewidth]{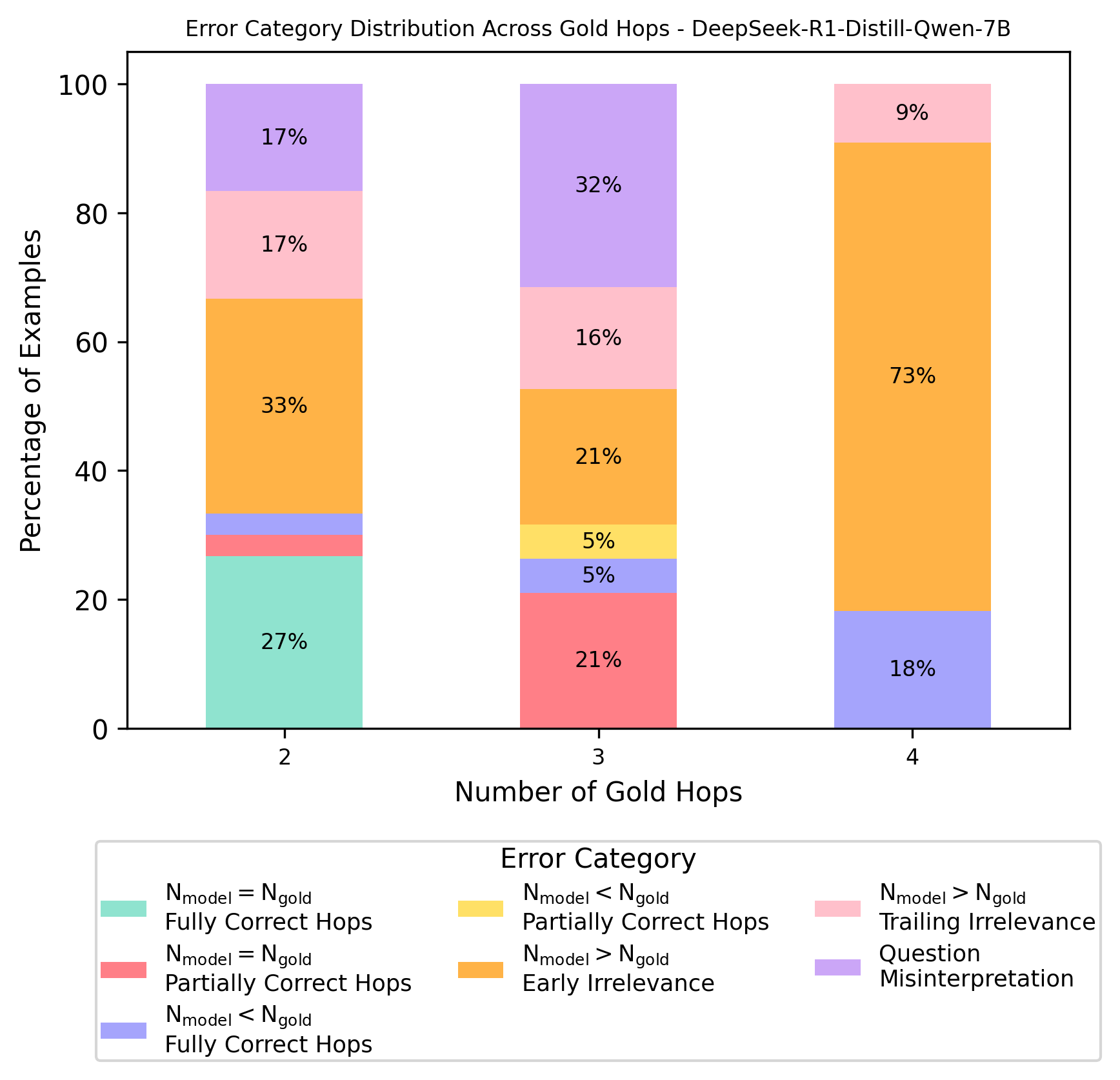}
        \caption{Qwen-7B (Distill)}
        \label{fig:qwen7b_hopwise}
    \end{subfigure}
    \hfill
    \begin{subfigure}[b]{0.48\textwidth}
        \centering
        \includegraphics[width=\linewidth]{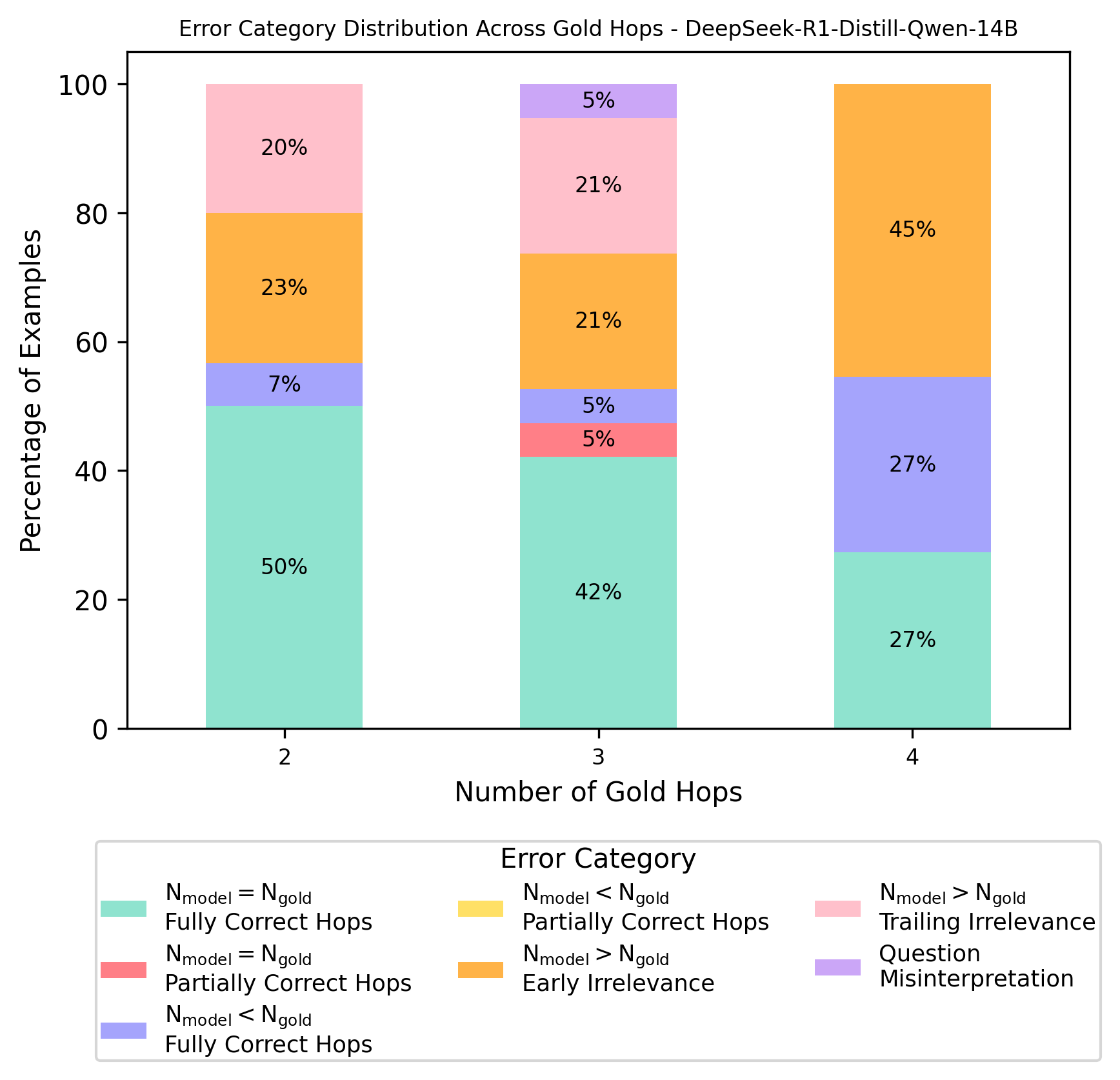}
        \caption{Qwen-14B (Distill)}
        \label{fig:qwen14b_hopwise}
    \end{subfigure}

    \vspace{1em}

    \begin{subfigure}[b]{0.48\textwidth}
        \centering
        \includegraphics[width=\linewidth]{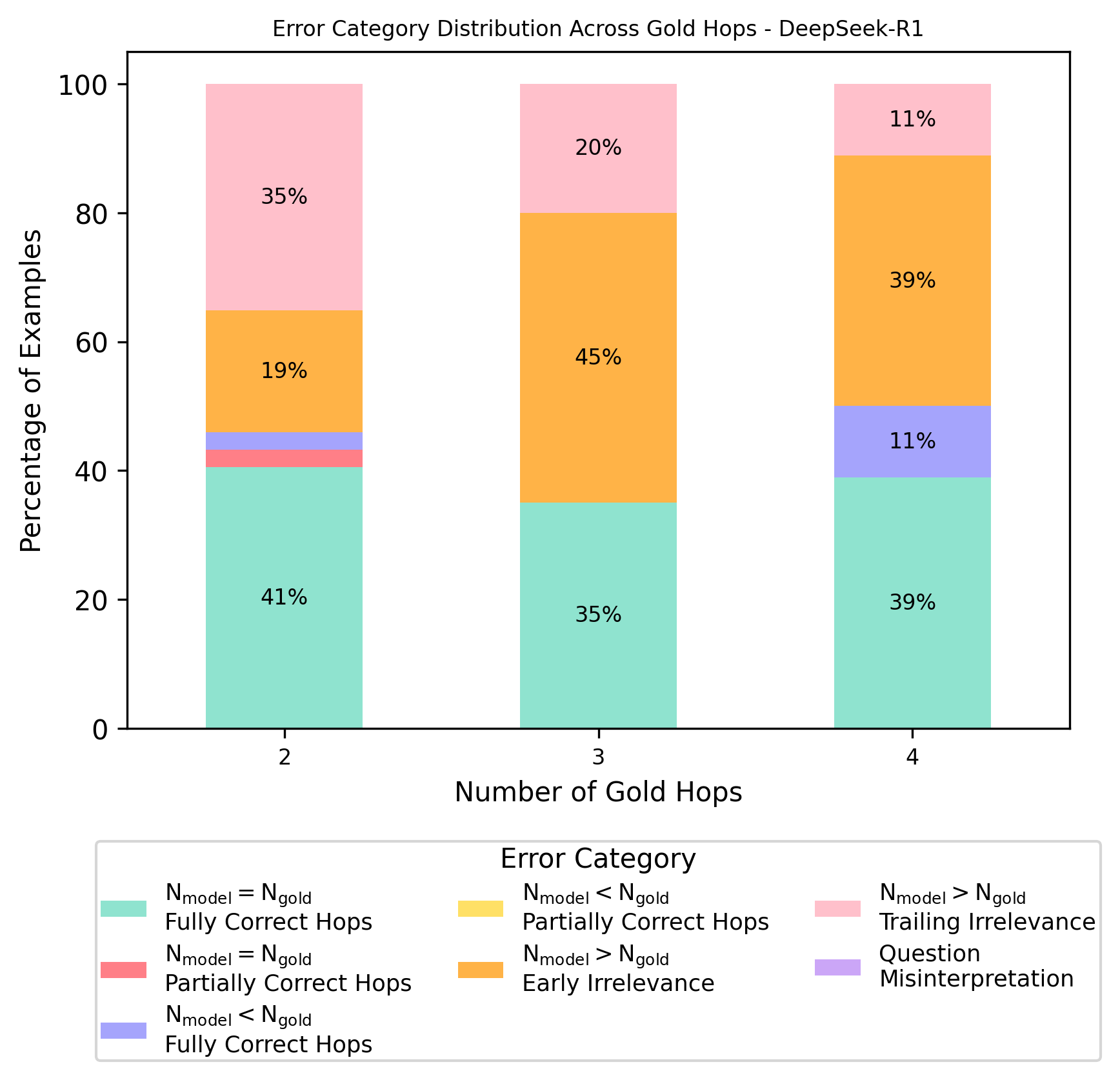}
        \caption{DeepSeek-R1}
        \label{fig:deepseekr1_hopwise}
    \end{subfigure}
    \hfill
    \begin{subfigure}[b]{0.48\textwidth}
        \centering
        \includegraphics[width=\linewidth]{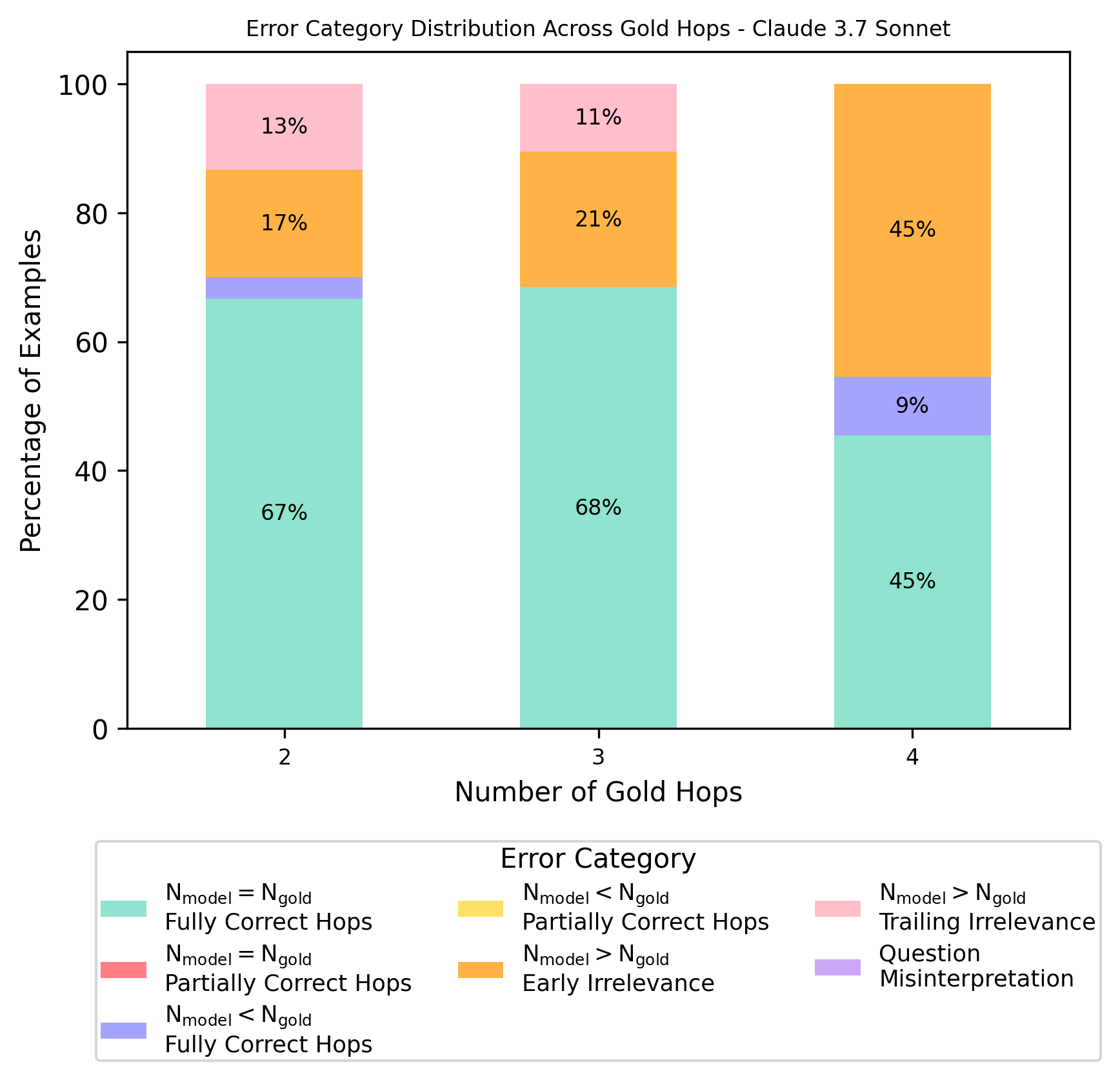}
        \caption{Claude 3 Sonnet}
        \label{fig:claude_hopwise}
    \end{subfigure}

    \caption{Hop-wise distribution of reasoning errors on MuSiQue for four models. Subplots (a)–(d) show how models vary in reasoning step correctness and overhopping behavior.}
    \label{fig:hopwise_all}
\end{figure*}

\end{document}